\documentclass[runningheads]{llncs}

\usepackage[mobile]{eccv}
\usepackage{eccvabbrv}

\usepackage{graphicx}
\usepackage{booktabs}
\usepackage{tabularx}
\usepackage{adjustbox}
\usepackage{siunitx}
\usepackage{subcaption}

\makeatletter
\newcommand{\printfnsymbol}[1]{%
        \textsuperscript{\@fnsymbol{#1}}%
}
\makeatother

\usepackage[accsupp]{axessibility}
\usepackage[pagebackref,breaklinks,colorlinks,citecolor=eccvblue]{hyperref}
\usepackage{orcidlink}

\begin{document}

\title{Exploring Multi-modal Neural Scene Representations With Applications \\ on Thermal Imaging} 

\titlerunning{Exploring Multi-modal Neural Scene Representations}

\author{Mert Özer\thanks{Authors contributed equally to this work.} \and
Maximilian Weiherer\printfnsymbol{1} \and
Martin Hundhausen \and
Bernhard Egger}

\authorrunning{M.~Özer et al.}

\institute{Friedrich-Alexander-Universität Erlangen-Nürnberg\\
\email{firstname.lastname@fau.de}}

\maketitle

\begin{abstract}
  Neural Radiance Fields (NeRFs) quickly evolved as the new de-facto standard for the task of novel view synthesis when trained on a set of RGB images.
  In this paper, we conduct a comprehensive evaluation of neural scene representations, such as NeRFs, in the context of multi-modal learning.
  Specifically, we present four different strategies of how to incorporate a second modality, other than RGB, into NeRFs: (1) training from scratch independently on both modalities; (2) pre-training on RGB and fine-tuning on the second modality; (3) adding a second branch; and (4) adding a separate component to predict (color) values of the additional modality.
  We chose thermal imaging as second modality since it strongly differs from RGB in terms of radiosity, making it challenging to integrate into neural scene representations.
  For the evaluation of the proposed strategies, we captured a new publicly available multi-view dataset, \textit{ThermalMix}, consisting of six common objects and about 360 RGB and thermal images in total.
  We employ cross-modality calibration prior to data capturing, leading to high-quality alignments between RGB and thermal images.
  Our findings reveal that adding a second branch to NeRF performs best for novel view synthesis on thermal images while also yielding compelling results on RGB.
  Finally, we also show that our analysis generalizes to other modalities, including near-infrared images and depth maps.
  Project page: \url{https://mert-o.github.io/ThermalNeRF/}.
  \keywords{Multi-modal Learning \and NeRF \and Thermal Imaging}
\end{abstract}

\section{Introduction}
Novel view synthesis pertains to the generation of new perspectives from an existing set of images.
Historically, this problem has been tackled using conventional techniques like structure-from-motion \cite{Schoenberger2016}, multi-view stereo \cite{Furukawa2015}, or image-based rendering techniques \cite{Shum2000}, and, recently, through the adoption of neural networks, prominently Neural Radiance Fields (NeRFs) \cite{Mildenhall2020}.
NeRFs offer a paradigm shift by encapsulating the scene within a continuous radiance field, allowing the representation of volume density and view-dependent RGB colors in a four-dimensional space.

\begin{figure}
     \centering
    \includegraphics[width=\linewidth]{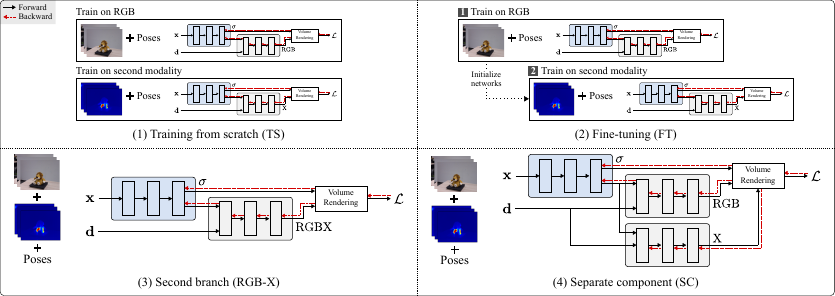}
     \caption{Overview of the four strategies that we compare within this work. In the first strategy (TS), we train a NeRF-like base model (Instant-NGP \cite{Mueller2022} in our case) from scratch, separately for RGB and the second modality. In the second strategy (FT), we first pre-train our base model on RGB data and then fine-tune on images of the second modality. While RGB-X adds a second branch, strategy four (SC) adds an extra \textit{network} to predict values of the additional modality. Note that RGB-X and SC yield a \textit{single}, multi-modal scene representation, whereas TS and FT always result in two separate models, one for each modality.}
     \label{fig:overview}
\end{figure}

On the other hand, multi-modal imaging, characterized by the simultaneous acquisition and processing of multiple data types from different \textit{optical sensors} (note that this definition explicitly excludes, \eg, text) has shown its significance across myriad applications, ranging from surface reconstruction \cite{Zollhöfer2018,Han2019,Koc2021} and image segmentation \cite{Meng2023} to applications in remote sensing \cite{Joshi2016,Wakeford2019} and medical imaging \cite{MoosaviTayebi2015,Petneházy2023}.
We believe that the integration of multi-modal information into modern neural scene representations like NeRFs could potentially enhance the depth and richness of the synthesized views, offering more detailed and nuanced scene reconstructions.
Indeed, this has been confirmed by a recent line of work, attempting to build multi-modal NeRFs from RGB and depth information \cite{Deng2021,Zhu2022,Herau2023,Xliu2023,zhang2023,Zhu2023,tao2024} as well as RGB and near-infrared images \cite{Poggi2022,Zhu2023}.

Through recent advancements, NeRFs have undergone transformative improvements \cite{gao2023}.
However, transitioning into a multi-modal environment introduces a host of complexities.
Notably, camera pose estimation for non-RGB images becomes a formidable challenge.
Established methods such as structure-from-motion can falter due to the unique features intrinsic to multi-modal (especially, multi-spectral) images.
Furthermore, the alignment of multi-sensory imagery necessitates a representation in a common coordinate system, requiring either offline cross-modality calibration or \textit{learning} of relative transformations between multiple sensors during training.
Ultimately, it is also unclear how to best integrate multi-modality into modern neural scene representations; a question, which is, to date, largely unexplored and that we will address in this work.

In this paper, we conduct a comprehensive evaluation of neural scene representations, specifically, NeRFs, within a multi-modal context.
We propose four different strategies of how to include a second modality (other than RGB) into a NeRF-like scene representation: (1) training from scratch, (2) fine-tuning, (3) adding a second branch, and (4) adding a separate component, see Fig. \ref{fig:overview}.
We chose thermal (\ie, far-infrared) imaging as the second modality for this work, since we consider modeling thermal images, among all existing imaging modalities, to be one of the hardest.
Compared to RGB images, thermal images are extremely feature-less (also, most of the features are blurry), and exhibit relatively low texture resolution even when captured with high-end cameras.
As we demonstrate in the supp. material, this causes serious problems in estimating (reliable) camera poses.
Moreover, because thermal images look so different compared to RGB, the underlying scene's geometry will differ from an RGB-derived geometry, rendering it non-trivial to design a neural scene representation that combines both modalities.
We evaluate the proposed strategies on a newly captured, object-centric dataset that includes multi-view RGB and thermal images of six common objects.
Our dataset, which we name \textit{ThermalMix} (because it is a mix of RGB and thermal images), consists of three forward-facing and three 360-degree scenes and comes with approximately 360 images in total.
As opposed to recent related works \cite{Poggi2022,Zhu2023}, we employ cross-modality calibration, yielding almost perfectly aligned RGB and thermal images.
Finally, we show that our results also generalize to other modalities, including near-infrared images, ultimately covering the whole bandwidth of the infrared spectrum within this work.

From a practical perspective, there is a huge number of potential domains and applications for which multi-modal neural scene representations integrating RGB \textit{and} thermal imagery could be of interest.
For instance, thermal imaging was successfully employed in agriculture \cite{vadivambal2011,ishimwe2014,khanal2017,messina2020}, medicine \cite{ring2012,aggarwal2023,deSouza2023,shaikh2019}, plant sciences \cite{jones2004,still2019,pineda2020}, aviation \cite{stumper2015}, defense systems \cite{akula2011}, food industry \cite{gowen2010}, and more (see also surveys \cite{gade2014,rai2017,wilson2023}).
Recently, a combination of RGB and thermal images was used for semantic segmentation \cite{shivakumar2020,huo2023,li2023}, defect detection \cite{yang2023}, traffic monitoring \cite{alldieck2016}, 3D reconstruction in medicine \cite{deSouza2023}, and food segmentation \cite{raju2023}.
Virtually all of the aforementioned applications (in which a combination of RGB and thermal data has proven to be beneficial) can be transferred into 3D, thus allowing for, \eg, better food segmentation in 3D scenes.

In summary, the core contributions of this paper are three-fold:
\begin{itemize}
\item We present a comprehensive study comparing four different strategies of how to learn multi-modal NeRFs based on RGB and thermal imagery.
\item We propose the first \textit{multi-view} dataset, named \textit{ThermalMix}, of high-quality aligned RGB and thermal images captured from six common objects.
\item We demonstrate that our results also generalize to other modalities, including near-infrared images and depth maps.
\end{itemize}
Our dataset is publicly available to foster future research and to serve as a benchmark, see \url{https://mert-o.github.io/ThermalNeRF/}.

\section{Related Work}
\textbf{Neural Radiance Fields (NeRFs).}
NeRF \cite{Mildenhall2020} utilizes a continuous function $F$ to characterize a scene, built upon simple, fully connected neural network layers.
Given a spatial coordinate $\mathbf{x}\in\mathbb{R}^3$ and its associated viewing direction $\mathbf{d}\in\mathbb{S}^2$, NeRF deduces both, the radiance $\mathbf{c}$ and density $\sigma$ as $F(\mathbf{x}, \mathbf{d}; \Theta) = (\boldsymbol{\mathrm{c}}, \sigma)$, where $\Theta$ stands for the parameters of the neural network.
For determining the final color of a pixel, ray marching is central.
Specifically, the accumulated color $\hat{\mathbf{c}}(\mathbf{r})$ of a camera ray $\mathbf{r}(t)=\mathbf{o}+t\mathbf{d}$ is computed using volume rendering:
\begin{equation}
    \hat{\mathbf{c}}(\mathbf{r}) = \sum_{i=1}^N T_i\left(1-\exp\left(-\sigma_i\delta_i\right)\right)\mathbf{c}_i,\quad\text{where}\quad T_i=\exp\left(-\sum_{j=1}^{i-1}\sigma_j\delta_j\right)
\end{equation}
and $\delta_i=t_{i+1}-t_i$.
Capturing high-frequency spatial variations in scenes hinges on positional encoding of spatial locations, $\mathbf{x}$ (and also viewing directions, $\mathbf{d}$).
The positional encoding introduced in NeRF relies on sinusoidal oscillations to scale frequencies logarithmically, thus enabling the neural network to attend to both, low and high-frequency details, see \cite{Mildenhall2020} and \cite{Tancik2020} for further information.

Numerous advancements have been made to improve upon the original NeRF framework, notably focusing on accelerating both the training and inference times from days to near real-time \cite{Deng2021,Hedman2021,Mueller2022,Yu2022,wang2023}.
One pivotal work in this regard is Instant-NGP \cite{Mueller2022}, which revises the architecture by partitioning the single unified multi-layer perception (MLP) responsible for coarse and dense sampling into two distinct networks dedicated to density and color estimation.
Additionally, Instant-NGP introduces \textit{Multiresolution Hash Encoding} for input data, enabling the use of smaller MLPs and thereby speeding up the training process without compromising the rendering quality.
In this paper, we adopt both, the architectural design and input encoding employed in Instant-NGP.
 
\textbf{Multi-modal NeRFs.}
Integrating multi-modality into NeRFs is a fairly new field of research, and only a few works exist that try to combine different modalities.
Most of the recent multi-modal NeRFs have been trained on RGB images and some kind of depth information, originating either from LiDAR scans \cite{Herau2023,zhang2023,Zhu2023,tao2024}, RGB-D images \cite{Deng2021,Zhu2022}, or ToF data \cite{Xliu2023}.
Moreover, few works recently tried to build multi-modal NeRFs from RGB and near-infrared images.
Based on computed camera poses, \cite{Zhu2023} first back-projects RGB and infrared images into 3D, yielding a coarse point cloud for both modalities, and then estimates relative transformations between sensors using point cloud registration.
Using RGB camera poses computed from COLMAP \cite{Schoenberger2016}, X-NeRF \cite{Poggi2022} \textit{learns} relative poses to the infrared sensor during training, and leverages \textit{Normalized Cross-Device Coordinates} to deal with different camera intrinsics.
Both groups do not share their code and/or rely on private datasets.
Lastly, we note concurrent work \cite{hassan2024,ye2024,xu2024,lin2024}, trying to incorporate thermal data into NeRF.

The majority of the methods mentioned above (including \cite{Poggi2022,Zhu2023}) implicitly assume a \textit{shared} volume density across different modalities, effectively using the architecture proposed in the third strategy (RGB-X) that we explore in this paper.
Furthermore, while previous works focus on how to simultaneously align \textit{and} fuse multi-modal images into a unified neural scene representation, the focus of this work is on the latter.
Essentially, we are interested in the following question: Assuming a multi-modal dataset with perfectly aligned images, what is the best way to integrate two imaging modalities into a neural scene representation?

\section{Method}
\label{sec:method}
We present four different strategies of how to include an additional modality, other than RGB, into neural scene representations: (1) training a base model from scratch, (2) fine-tuning, (3) adding a second branch to the base model, and (4) adding an extra component for the second modality.
Notably, while the last two strategies result in a \textit{single}, multi-modal scene representation, the first two strategies always yield two separate models: one for RGB, and one for the second modality.
A schematic overview of the four strategies is given in Fig. \ref{fig:overview}.

Throughout this work, we use Instant-NGP \cite{Mueller2022} as our base model.
In brief, Instant-NGP's architecture comprises a density network composed of three fully-connected layers with 64 hidden dimensions each, as well as a color network featuring three fully-connected layers but with 32 hidden dimensions each.
The density network takes hash-encoded coordinates as input and outputs a 16-dimensional vector, providing point-wise densities along with a 15-dimensional geometric descriptor.
The color network processes the 15-dimensional descriptor in conjunction with an encoded viewing direction to generate view-dependent RGB color values.
It is noteworthy that the density network's configuration remains consistent across all four strategies explained in the following.

\textbf{Training from scratch (TS).}
In the first and simplest strategy, we train our base model from scratch, separately for RGB and the second modality.
Since it is extremely challenging to compute reliable camera poses for thermal images using standard structure-from-motion approaches (see supp. material), we leverage poses derived from the corresponding RGB images and employ the following NeRF-like loss for training on thermal images:
\begin{equation}
    \mathcal{L}_t=\sum_{\mathbf{r}\in\mathcal{R}}\left(\hat{t}(\mathbf{r})-t(\mathbf{r})\right)^2,
    \label{eq:thermal_loss}
\end{equation}
where $\mathcal{R}$ is a set of rays, and $\hat{t}(\mathbf{r})$ and $t(\mathbf{r})$ are the predicted and ground-truth temperature values, respectively.
Note that this loss is task-specific and may vary depending on the modality at hand.
For training on RGB images, we use the standard rendering loss between predicted and true pixel color, $\hat{\mathbf{c}}(\mathbf{r})$ and $\mathbf{c}(\mathbf{r})$, respectively:
\begin{equation}
    \mathcal{L}_c=\sum_{\mathbf{r}\in\mathcal{R}}\Vert\hat{\mathbf{c}}(\mathbf{r})-\mathbf{c}(\mathbf{r})\Vert_2^2.
    \label{eq:color_loss}
\end{equation}
This strategy serves as our baseline.

\textbf{Fine tuning (FT).}
Our second strategy first trains the base model on RGB images and then fine-tunes on images from the second modality, for which we again leverage RGB-derived camera poses.
The idea behind this strategy is based on the assumption that the underlying scene's geometry is similar in both modalities and that training of the second modality can profit from being initialized with RGB data.
For training on RGB images, we apply the same loss as in (\ref{eq:color_loss}); similarly, fine-tuning on thermal images is done using the loss in (\ref{eq:thermal_loss}).

\begin{figure}
\captionsetup[subfigure]{labelformat=empty}
     \centering
     \begin{subfigure}[b]{0.16\textwidth}
         \centering
         \includegraphics[width=\textwidth]{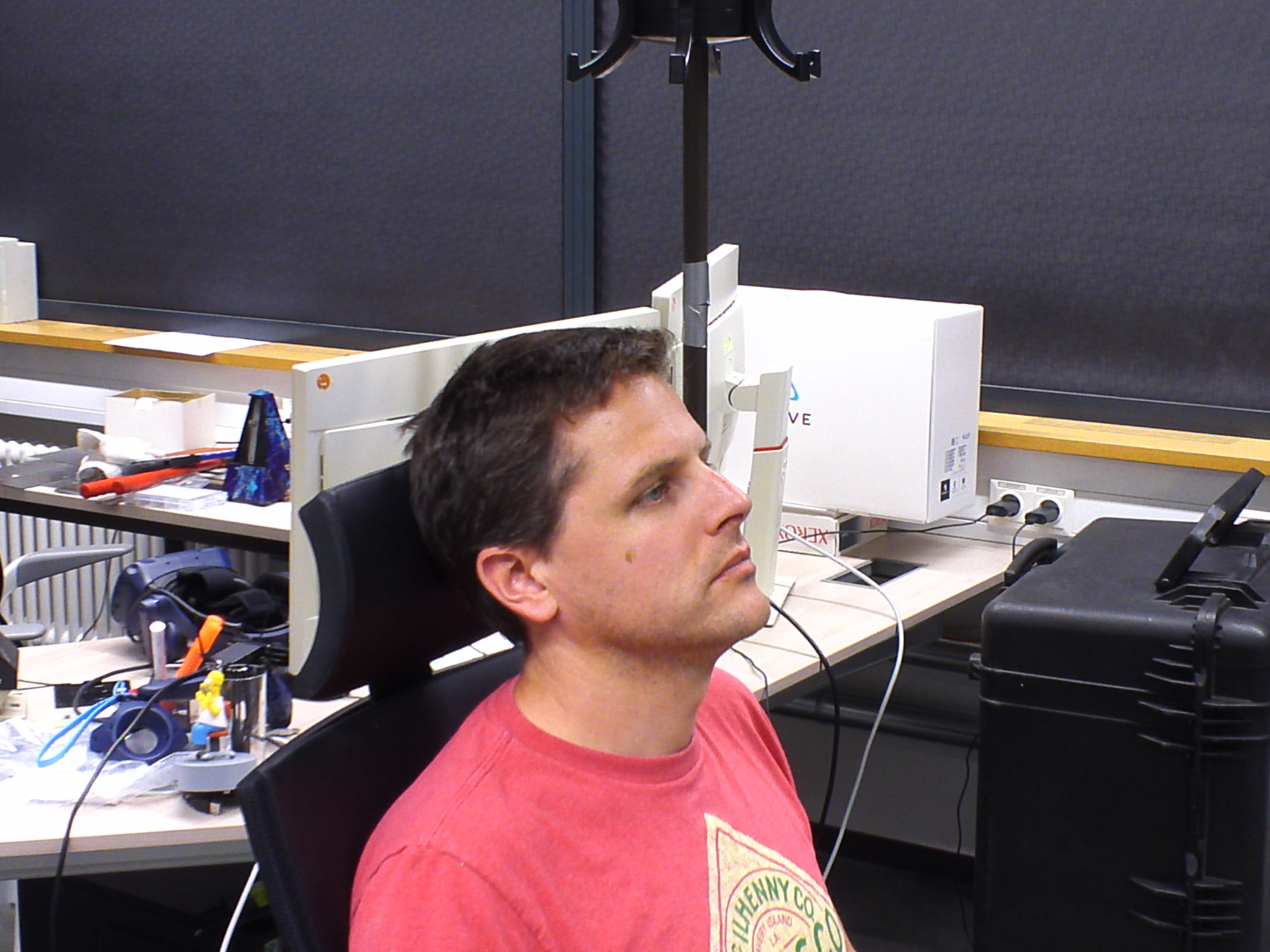}
     \end{subfigure}
     \begin{subfigure}[b]{0.16\textwidth}
         \centering   
         \includegraphics[width=\textwidth]{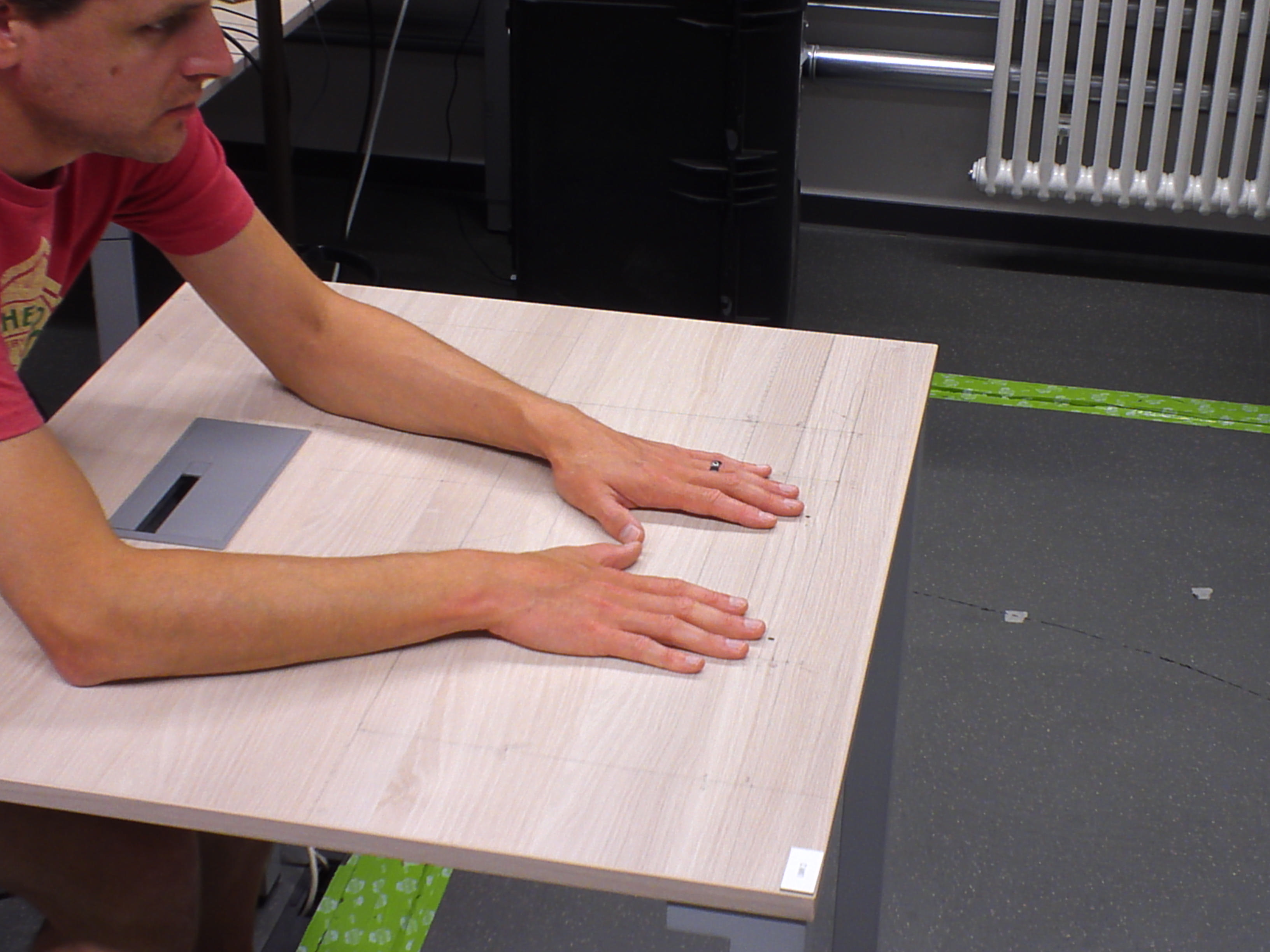}
     \end{subfigure}
     \begin{subfigure}[b]{0.16\textwidth}
         \centering
         \includegraphics[width=\textwidth]{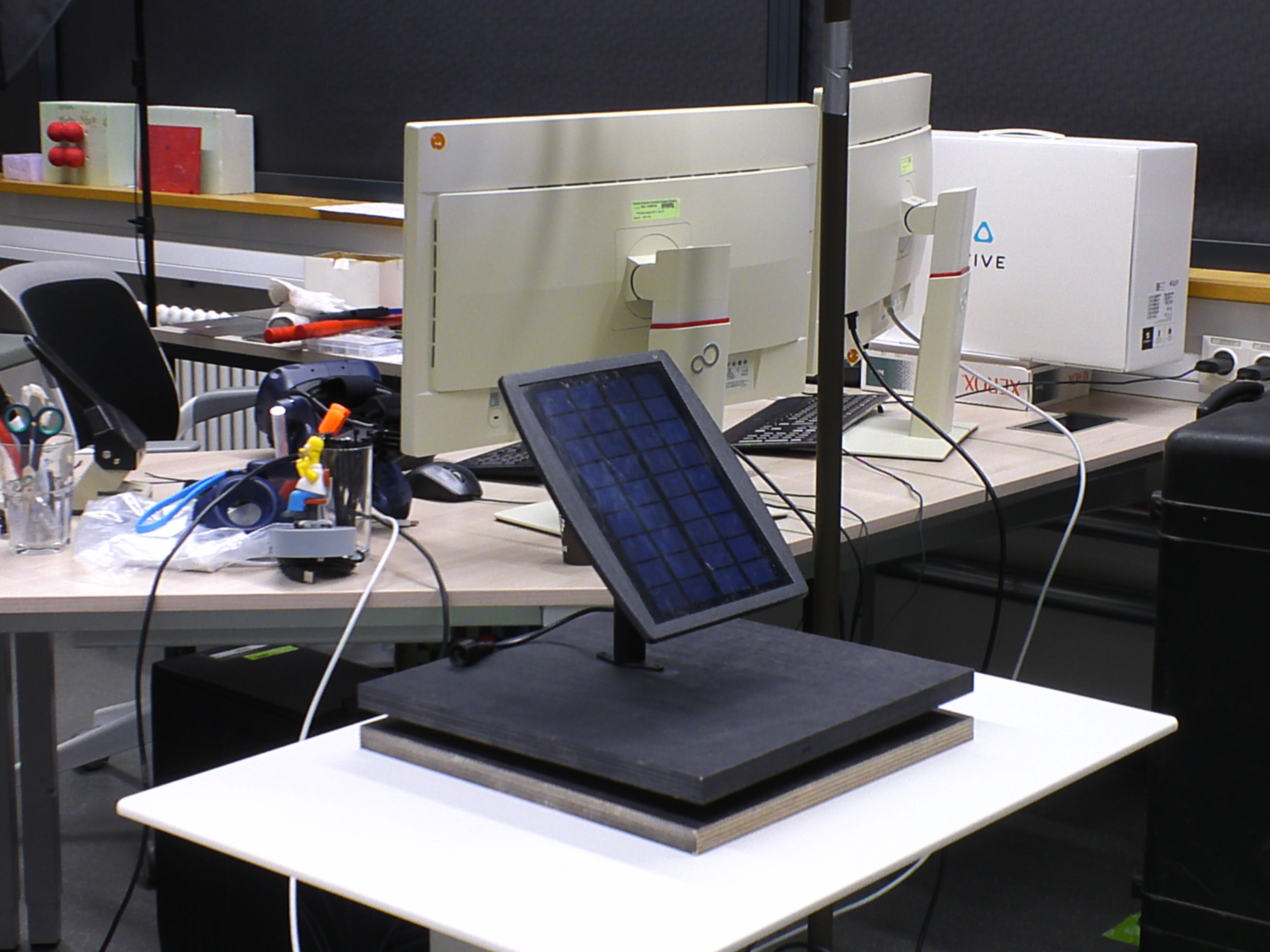}
     \end{subfigure}
     \begin{subfigure}[b]{0.16\textwidth}
         \centering
         \includegraphics[width=\textwidth]{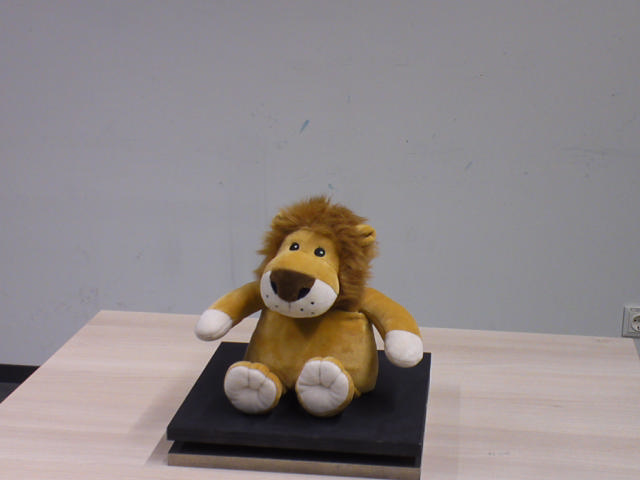}
     \end{subfigure}
     \begin{subfigure}[b]{0.16\textwidth}
         \centering
         \includegraphics[width=\textwidth]{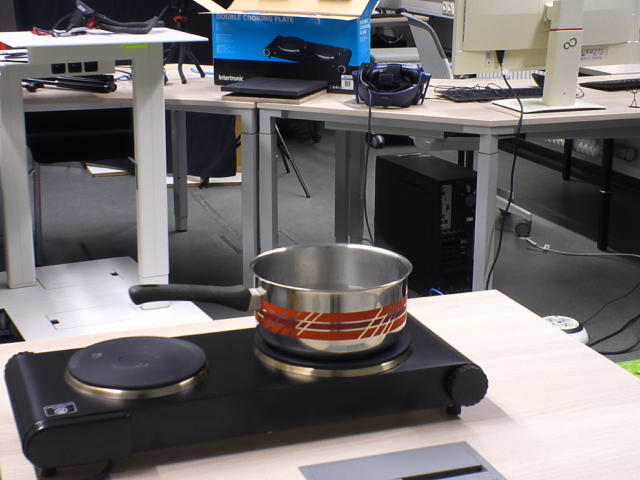}
     \end{subfigure}
     \begin{subfigure}[b]{0.16\textwidth}
         \centering
         \includegraphics[width=\textwidth]{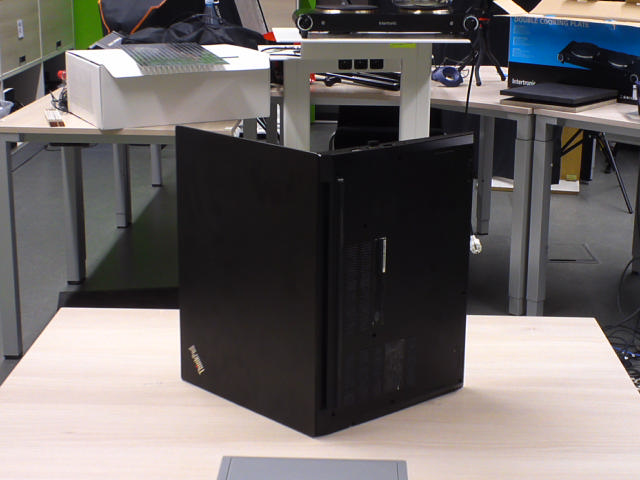}
     \end{subfigure}
     
    \vspace{0.05cm} 
     \begin{subfigure}[b]{0.16\textwidth}
         \centering
         \includegraphics[width=\textwidth]{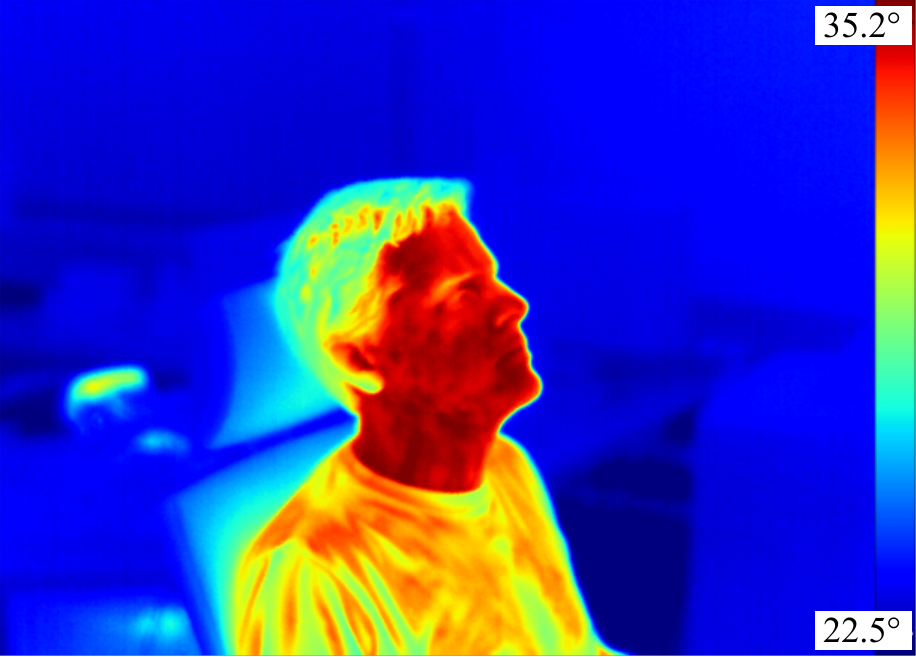}
         \caption{\textsc{Face}}
     \end{subfigure}
     \begin{subfigure}[b]{0.16\textwidth}
         \centering
         \includegraphics[width=\textwidth]{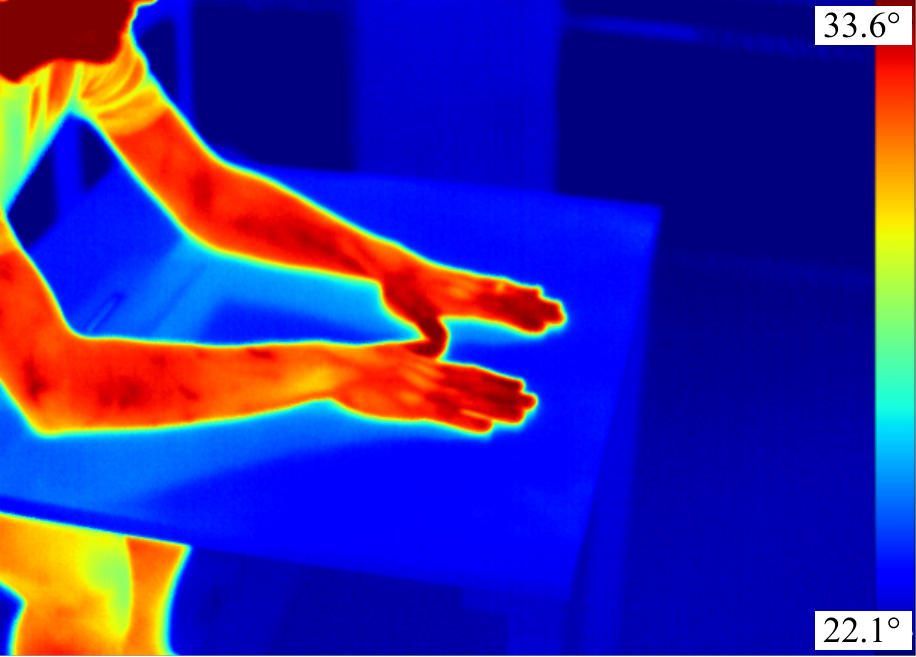}
         \caption{\textsc{Hand}}
     \end{subfigure}
     \begin{subfigure}[b]{0.16\textwidth}
         \centering
         \includegraphics[width=\textwidth]{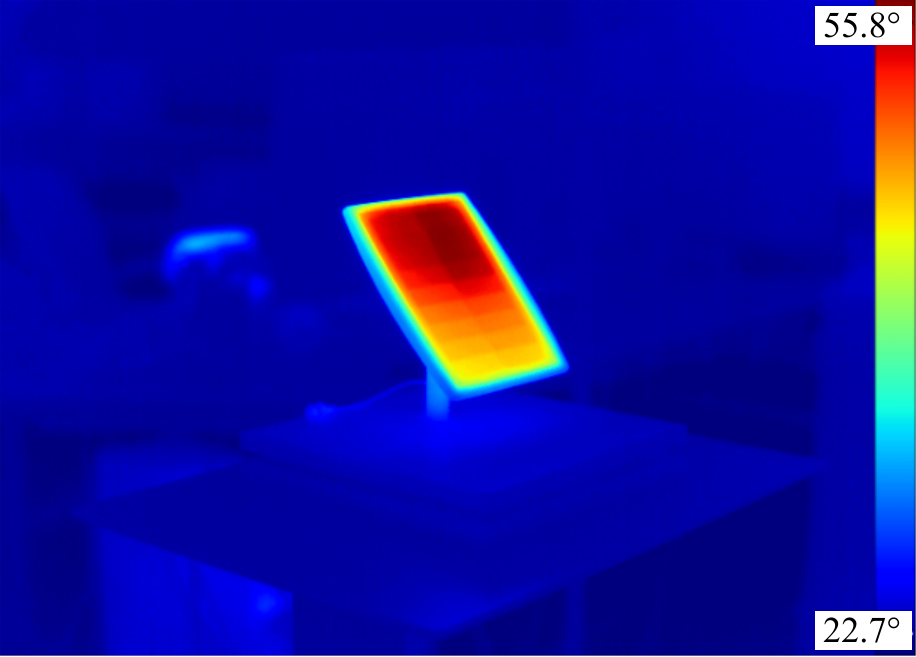}
         \caption{\textsc{Panel}}
     \end{subfigure}
     \begin{subfigure}[b]{0.16\textwidth}
         \centering
         \includegraphics[width=\textwidth]{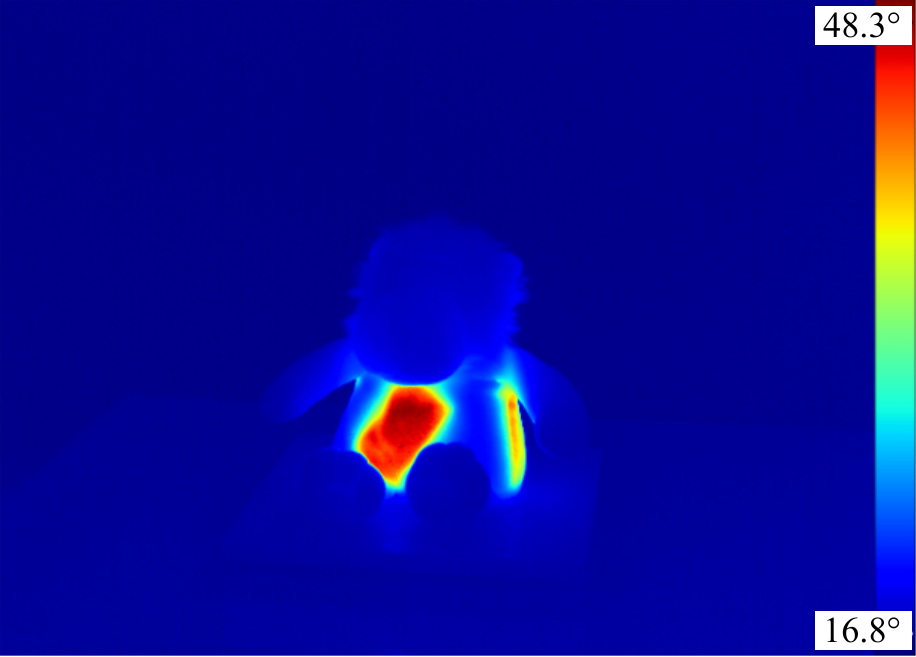}
         \caption{\textsc{Lion}}
     \end{subfigure}
     \begin{subfigure}[b]{0.16\textwidth}
         \centering
         \includegraphics[width=\textwidth]{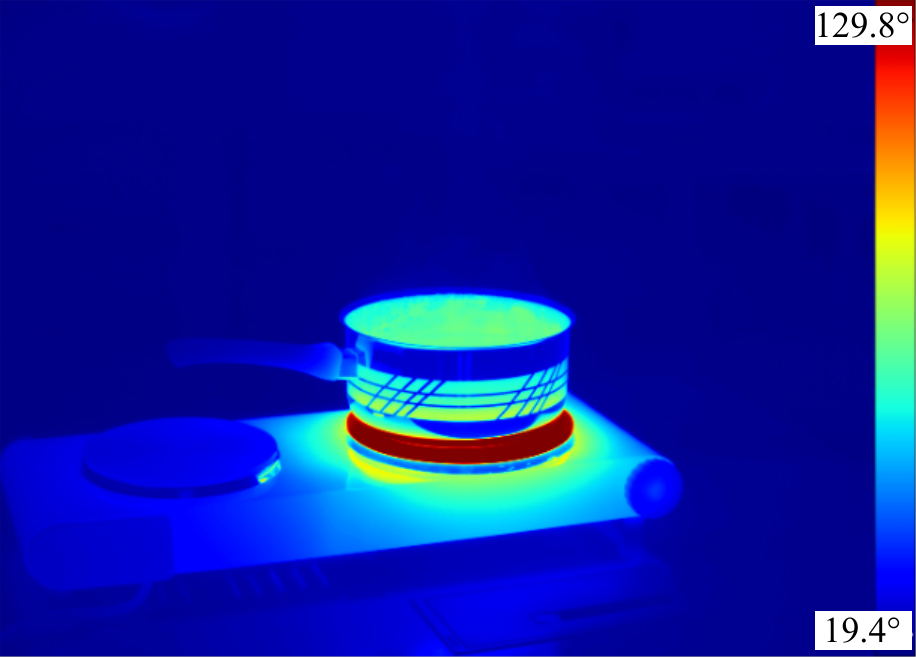}
         \caption{\textsc{Pan}}
     \end{subfigure}
     \begin{subfigure}[b]{0.16\textwidth}
         \centering
         \includegraphics[width=\textwidth]{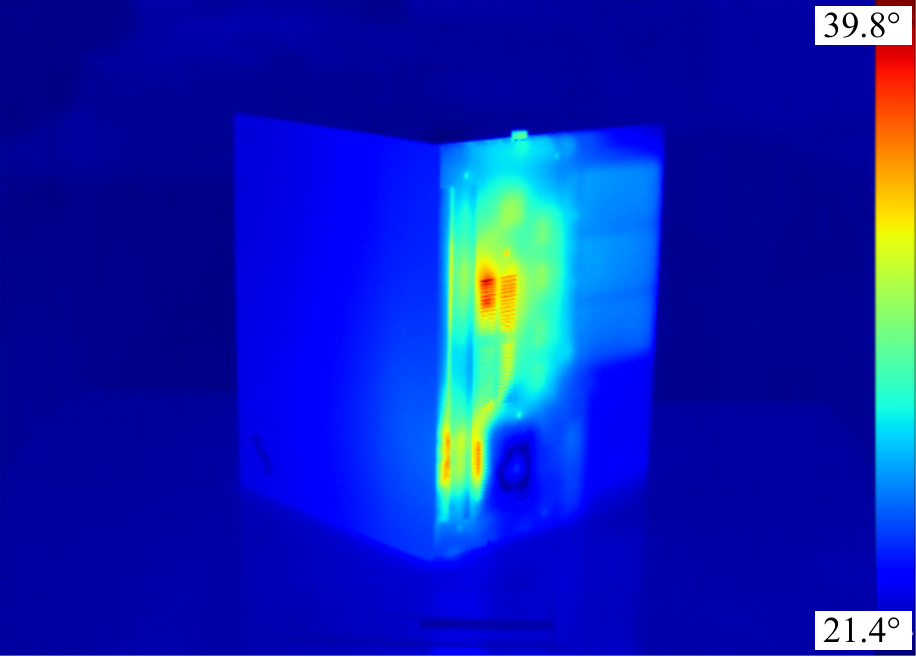}
         \caption{\textsc{Laptop}}
     \end{subfigure}
     \caption{Overview of our newly-captured dataset containing high-quality aligned RGB and thermal images of six common objects. \textsc{Face}, \textsc{Hand}, and \textsc{Panel} are forward-facing scenes consisting of around 40 images each. \textsc{Lion}, \textsc{Pan}, and \textsc{Laptop} are 360-degree scenes, where each scene has around 80 images.}
     \label{fig:datasets}
\end{figure}

\textbf{Second branch (RGB-X).}
Our third strategy leverages multi-task learning by adding a second branch to the color network that predicts values of the second modality.
Although sharing a similar motivation as FT, RGB-X is the first of two proposed strategies that incorporate the additional modality into a single model.
During training, we back-propagate both, RGB \textit{and} predicted values of the second modality through the density network.
Consequently, the density network is not only influenced by RGB but also the second modality.
In our case, the color network is adapted to produce a four-dimensional RGB$t$ output, comprising RGB color values \textit{and} temperatures, $t$.

\textbf{Separate component (SC).}
Contrary to utilizing a shared density for both modalities as in RGB-X, there are scenarios where this approach is sub-optimal for integrating diverse modalities.
For instance, in our case, when using thermal and RGB imagery, the object's geometry under infrared deviates from its visible representation, please see supp. material.
To rectify these disparities, and hence being able to predict accurate temperature values, especially for visible regions, it becomes imperative to leverage \textit{only} RGB-derived densities.
To account for this, our last strategy adds an extra component to the base model that solely predicts values of the second modality, but, contrary to RGB-X, \textit{restricts} back-propagation to the density network during training.
This constraint encourages the additional network to implicitly approximate RGB-derived geometry and prevents densities from being influenced by the second modality.
We use a weighted combination of the previous loss functions from (\ref{eq:thermal_loss}) and (\ref{eq:color_loss}) for training on thermal images:
\begin{equation}
    \mathcal{L}=\omega_c\mathcal{L}_c+\omega_t\mathcal{L}_t,
    \label{eq:combined_loss}
\end{equation}
where we keep $w_c=w_t=1$ constant (see supp. material for an ablation).
The separate network predicting temperatures shares the same architecture as our base model's color network.

\section{Dataset}
\label{sec:dataset}
We use a custom dataset containing RGB and thermal images of three forward-facing and three 360-degree scenes to compare previously explained strategies, see Fig. \ref{fig:datasets}.
In total, our dataset, which we call \textit{ThermalMix}, contains six common objects (\textsc{Face}, \textsc{Hand}, \textsc{Panel}, \textsc{Lion}, \textsc{Pan}, and \textsc{Laptop}) and is publicly available.

The data acquisition setup comprises a thermal camera (VarioCam HD, InfraTec GmbH, Germany) equipped with a $640\times 480$ pixel resolution at 60 Hz for both, RGB and infrared sensor, a measurable temperature range of $-40$ to $+2,000$ degrees Celsius with an accuracy of $\pm 1$ degree Celsius, and an infrared spectrum of 7.5 to \SI{14}{\micro\metre}.
Each of the six objects is placed on a table while the camera is moving around the object with a constant distance of about \SI{1}{\metre}.
Each forward-facing scene contains about 40 images, whereas about 80 images were taken for 360-degree scenes.

\textbf{Cross-modality calibration.}
Since RGB and thermal images are captured from different sensors each one of them having its own coordinate system, a calibration object is positioned at the scene's center prior to data capturing, based on which we estimate the relative transformation between both sensors.
Aligning images from different modalities is crucial in multi-modal reconstruction \cite{Poggi2022,Zhu2023}, and we are in need of a calibration target with easily identifiable features across both modalities.
Finding the correct calibration object for RGB and thermal imagery, however, is a non-trivial task itself, see, \eg, \cite{shivakumar2020} or \cite{swamidoss2021}.
As the two modalities strongly differ in radiosity, common materials, objects, and patterns (\eg, the classical checkerboard pattern printed on paper) can not be used, simply because they will not be visible in the infrared spectrum.
Instead, we chose a perforated plate made out of aluminum as a calibration target, see supp. material.
After slightly warming up (or cooling down) the plate, the holes are easily recognizable under both modalities, and we ultimately detect midpoints as matching features.
Finally, since the distance between the camera and the object is fixed, we derive camera poses for thermal images from the previously computed transformation and absolute poses estimated from the corresponding RGB images (using COLMAP \cite{Schoenberger2016}). 
Please see supp. material for more details.

\section{Results}
Based on our \textit{ThermalMix} dataset, we conducted extensive experiments to compare the proposed strategies.
We start by quantitatively and qualitatively evaluating their ability to reconstruct (during training left out) thermal images in Section \ref{subsec:results_thermal} and continue with presenting reconstruction results on RGB images in Section \ref{subsec:results_rgb}.
Finally, we also investigate if our results generalize to other modalities, including near-infrared images and depth maps, in Section \ref{subsec:results_other}.

\subsection{Implementation Details}
\textbf{Pre-processing.} 
For pre-processing, RGB and thermal images are normalized within the range $[0,1]$.
Notably, for thermal images, normalization is performed relative to the \textit{scene's} maximum temperature (\ie, the maximum temperature over all images of a scene), as opposed to the conventional per-image maximum.
This is necessary to ensure a consistent re-mapping after training.

\begin{table}
 \caption{Quantitative results on thermal images for (a) the three forward-facing scenes and (b) the three 360-degree scenes, measured using PSNR and SSIM (higher is better). Results were obtained from NeRFs trained on RGB and thermal data.}
    \begin{subtable}{.48\linewidth}
      \centering
        \begin{adjustbox}{width=\linewidth}
        \begin{tabular}{lcc|cc|cc|cc}
            \toprule
             & \multicolumn{2}{c|}{TS} & \multicolumn{2}{c|}{FT} & \multicolumn{2}{c|}{RGB-$t$} & \multicolumn{2}{c}{SC} \\
             \cmidrule(lr){2-3} \cmidrule(lr){4-5} \cmidrule(lr){6-7} \cmidrule(lr){8-9}
             & PSNR $\uparrow$ & SSIM $\uparrow$ & PSNR $\uparrow$ & SSIM $\uparrow$ & PSNR $\uparrow$ & SSIM $\uparrow$ & PSNR $\uparrow$ & SSIM $\uparrow$ \\
            \midrule
            \textsc{Face} & 30.34 & \textbf{0.77} & 30.04 & 0.75 & \textbf{33.44} & 0.68 & 32.10 & 0.66 \\
            \textsc{Hand} & 35.54 & \textbf{0.81} & 33.99 & 0.73 & \textbf{36.34} & 0.73 & 33.56 & 0.60 \\
            \textsc{Panel} & 31.21 & \textbf{0.74} & 29.66 & 0.55 & \textbf{31.36} & 0.61 & 27.31 & 0.38 \\
            \midrule\midrule
            & 32.36 & \textbf{0.77} & 31.23 & 0.68 & \textbf{33.71} & 0.67 & 30.99 & 0.55 \\
            \bottomrule
        \end{tabular}
    \end{adjustbox}
    \caption{Forward-facing scenes.}
    \end{subtable}\hfill
    \begin{subtable}{.48\linewidth}
      \centering
        \begin{adjustbox}{width=1.02\linewidth}
        \begin{tabular}{lcc|cc|cc|cc}
            \toprule
             & \multicolumn{2}{c|}{TS} & \multicolumn{2}{c|}{FT} & \multicolumn{2}{c|}{RGB-$t$} & \multicolumn{2}{c}{SC} \\
             \cmidrule(lr){2-3} \cmidrule(lr){4-5} \cmidrule(lr){6-7} \cmidrule(lr){8-9}
             & PSNR $\uparrow$ & SSIM $\uparrow$ & PSNR $\uparrow$ & SSIM $\uparrow$ & PSNR $\uparrow$ & SSIM $\uparrow$ & PSNR $\uparrow$ & SSIM $\uparrow$ \\
            \midrule
            \textsc{Lion} & 21.83 & 0.51 & 25.13 & 0.52 & \textbf{27.82} & \textbf{0.61} & 27.59 & 0.60 \\
            \textsc{Pan} & 20.46 & 0.53 & 24.14 & 0.50 & \textbf{27.48} & \textbf{0.54} & 26.43 & 0.53 \\
            \textsc{Laptop} & 23.15 & 0.37 & 24.95 & 0.49 & \textbf{30.17} & \textbf{0.59} & 28.07 & 0.53 \\
            \midrule\midrule
            & 21.81 & 0.47 & 24.74 & 0.50 & \textbf{28.49} & \textbf{0.58} & 27.36 & 0.55 \\
            \bottomrule
        \end{tabular}
    \end{adjustbox}
    \caption{360-degree scenes.}
    \end{subtable} 
    \label{tab:results_thermal}
\end{table}

\begin{figure}[t]
    \centering
    \includegraphics[width=\textwidth]{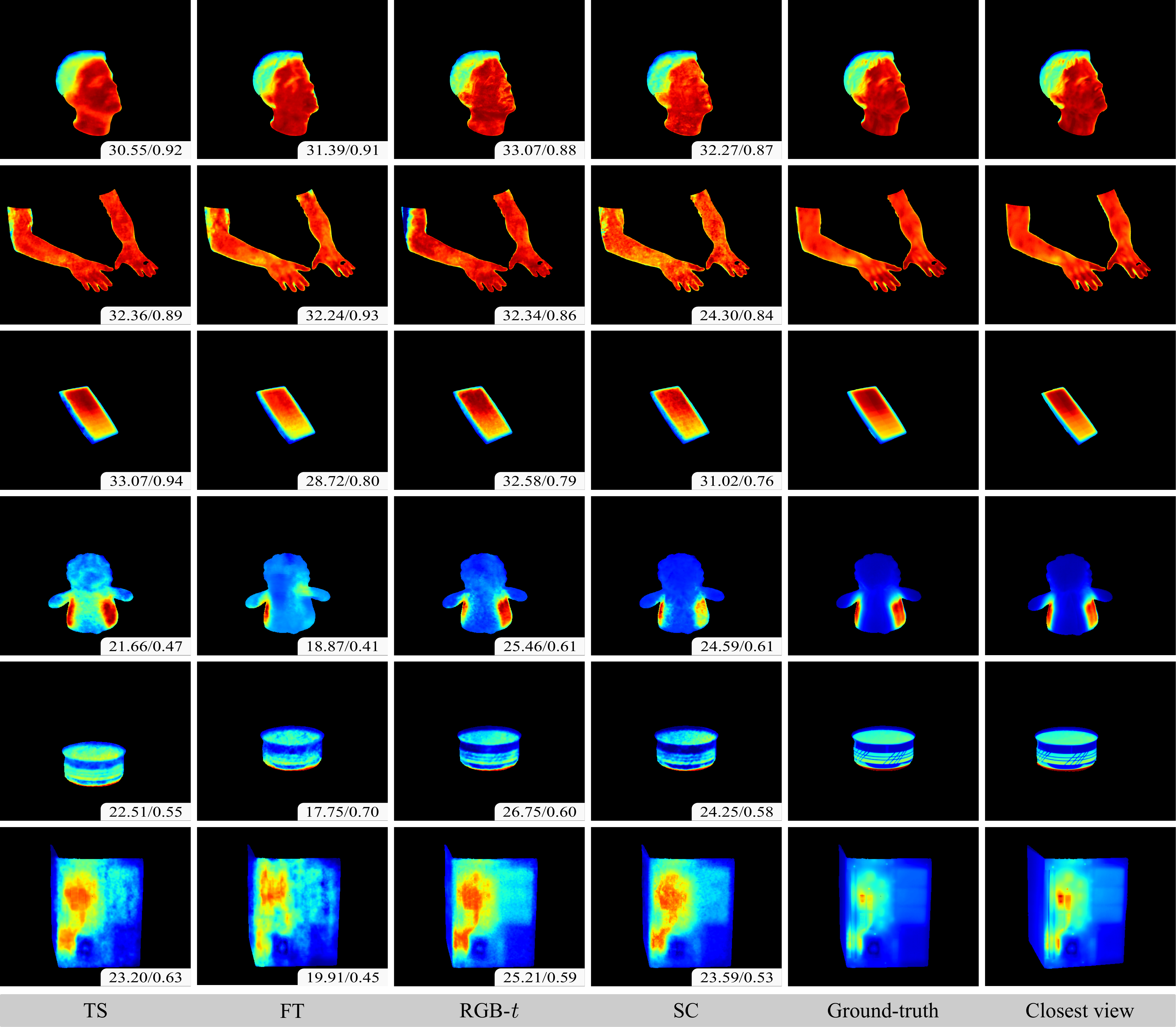}
    \caption{Reconstructions of a (left-out) thermal image from multi-modal neural scene representations trained on RGB and thermal data, arising from the four strategies that we compare. For each view, we also report PSNR and SSIM (higher is better). \textit{Closest view} denotes the nearest image in the training set.}
    \label{fig:results_thermal}
\end{figure}

\textbf{Training.} We use the same empirically determined parameters across all objects and scenes.
For FT, we pre-train for 6,000 iterations on RGB and fine-tune for another 4,000 iterations on thermal images.
Networks of the remaining strategies (TS, RGB-X, and SC) are trained for 10,000 iterations each.
Moreover, we used a batch of 4,096 rays per iteration, Adam \cite{Kingma2014} optimizer with default parameters, and a learning rate of 0.01.

\textbf{Evaluation.}
Following the literature, we report Peak Signal-to-Noise Ratio (PSNR) and the Structural Similarity Index Measure (SSIM) to evaluate our models, computed using leave-one-out cross-validation.
Specifically, in each training process, a single image is set aside as a test sample while the model is trained on the remaining data.
This procedure is repeated 10 times, and results were averaged.
Finally, since our primary focus remains on the temperatures of the scene's central object, we segment objects within test images and compute evaluation metrics only in regions covered by an object (for training, however, we use the full, unsegmented images).
Segmentation of thermal images is guided by RGB-derived segmentation masks.

\subsection{Evaluation on Thermal Images}
\label{subsec:results_thermal}
We present reconstruction results on thermal images separately for forward-facing and 360-degree scenes.

\textbf{Forward-facing scenes.}
Results for forward-facing scenes (\textsc{Face}, \textsc{Hand}, and \textsc{Panel}) are shown in Table \ref{tab:results_thermal}(a) and Fig. \ref{fig:results_thermal}.
As seen, RGB-X (which we denote as RGB-$t$ in the following) outperforms all other strategies when considering PSNR.
In terms of SSIM, however, TS performs best.
This outcome is not surprising, considering the fact that TS solely relies on thermal measurements for its evaluations.
More interestingly, we observe that RGB-$t$'s performance in terms of SSIM is closest to the performance of FT (absolute deviation of 0.13; second-best is 0.29 between TS and FT).
Since densities in RGB-$t$ are affected by back-propagated temperature values, this suggests that the RGB-$t$ density network somehow balances between RGB and thermal densities, allowing FT to further lean on the thermal densities during fine-tuning.
Furthermore, our empirical evaluation shows that the performance of SC seems to be inconsistent, varying with the object being reconstructed.
This variability is understandable given that SC does not integrate thermal densities, making the complexity of the scene a critical factor in the performance of the separate thermal network. 

\textbf{360-degree scenes.}
We report PSNRs and SSIMs for 360-degree scenes (\textsc{Lion}, \textsc{Pan}, and \textsc{Laptop}) in Table \ref{tab:results_thermal}(b) and qualitative results in Fig. \ref{fig:results_thermal}.
Contrary to forward-facing scenes, RGB-$t$ outperforms other strategies in both, PSNR \textit{and} SSIM for 360-degree scenes.
Also, another key observation is the minimal variation in both metrics across different objects, unlike other strategies.
This uniformity demonstrates the robustness of RGB-$t$ against the challenges inherent in thermal imagery and underscores the significance of using RGB images to guide thermal reconstruction.

Moreover, we observe that TS and FT perform poorly with regard to PSNR, showing that these strategies exhibit serious difficulties in achieving high-quality results on thermal images within a 360-degree context. 
A possible explanation for this effect may be the static nature of the background in thermal images, which tends to not have as much variation as seen in RGB images.
Ultimately, this introduces ambiguities, effectively hampering the network to distinguish between forward- and backward-facing views (this becomes even worse if the object itself is rotationally symmetrical, such as the pan, for instance).
Finally, we would like to note that TS, unlike FT, seems to be more sensitive to floaters.
Please refer to the supp. material for more information.

The performance of SC consistently ranks second to RGB-$t$ (both, for PSNR and SSIM), reinforcing the notion that relying solely on thermal images in 360-degree scenes can yield subpar results, primarily due to the near-static background.
In SC, where only the densities from the RGB data are utilized, we observe an improvement over TS but results are either comparable or inferior to RGB-$t$.
This outcome highlights the limitations of using only RGB densities in a thermal context and underlines the added value of incorporating RGB information directly, as done in the RGB-$t$.

\begin{table}[t]
\caption{Quantitative results on RGB images, measured using PSNR and SSIM (higher is better). Results were obtained from NeRFs trained on RGB and thermal data. FT is left out since its RGB component is similar to TS.}
\centering
 \begin{adjustbox}{width=0.5\linewidth}
        \begin{tabular}{lcc|cc|cc}
            \toprule
             & \multicolumn{2}{c|}{TS} & \multicolumn{2}{c|}{RGB-$t$} & \multicolumn{2}{c}{SC} \\
             \cmidrule(lr){2-3} \cmidrule(lr){4-5} \cmidrule(lr){6-7}
             & PSNR $\uparrow$ & SSIM $\uparrow$ & PSNR $\uparrow$ & SSIM $\uparrow$ & PSNR $\uparrow$ & SSIM $\uparrow$ \\
            \midrule
            \textsc{Face} & \textbf{30.78} & \textbf{0.85} & 29.46 & 0.79 & 30.12 & 0.82 \\
            \textsc{Hand} & \textbf{30.81} & \textbf{0.93} & 30.37 & 0.82 & 30.62 & 0.88 \\
            \textsc{Panel} & \textbf{30.56} & \textbf{0.84} & 29.81 &  0.79 & 30.03 & 0.80 \\
            \midrule
            \textsc{Lion} & \textbf{30.71} & \textbf{0.82} & 29.08 & 0.73 & 30.18 & 0.77 \\
            \textsc{Pan} & \textbf{29.59} & \textbf{0.76} & 29.33 & 0.73 & 29.40 & 0.72 \\
            \textsc{Laptop} & \textbf{29.70} & \textbf{0.74} & 29.42 & 0.72 & 29.45 & 0.72 \\
            \midrule\midrule
             & \textbf{30.36} & \textbf{0.82} & 29.58 & 0.76 & 29.67 & 0.79 \\
            \bottomrule
        \end{tabular}
    \end{adjustbox}
    \label{tab:uni_vs_multi}
\end{table}

\begin{figure}
    \centering
    \includegraphics[width=\textwidth]{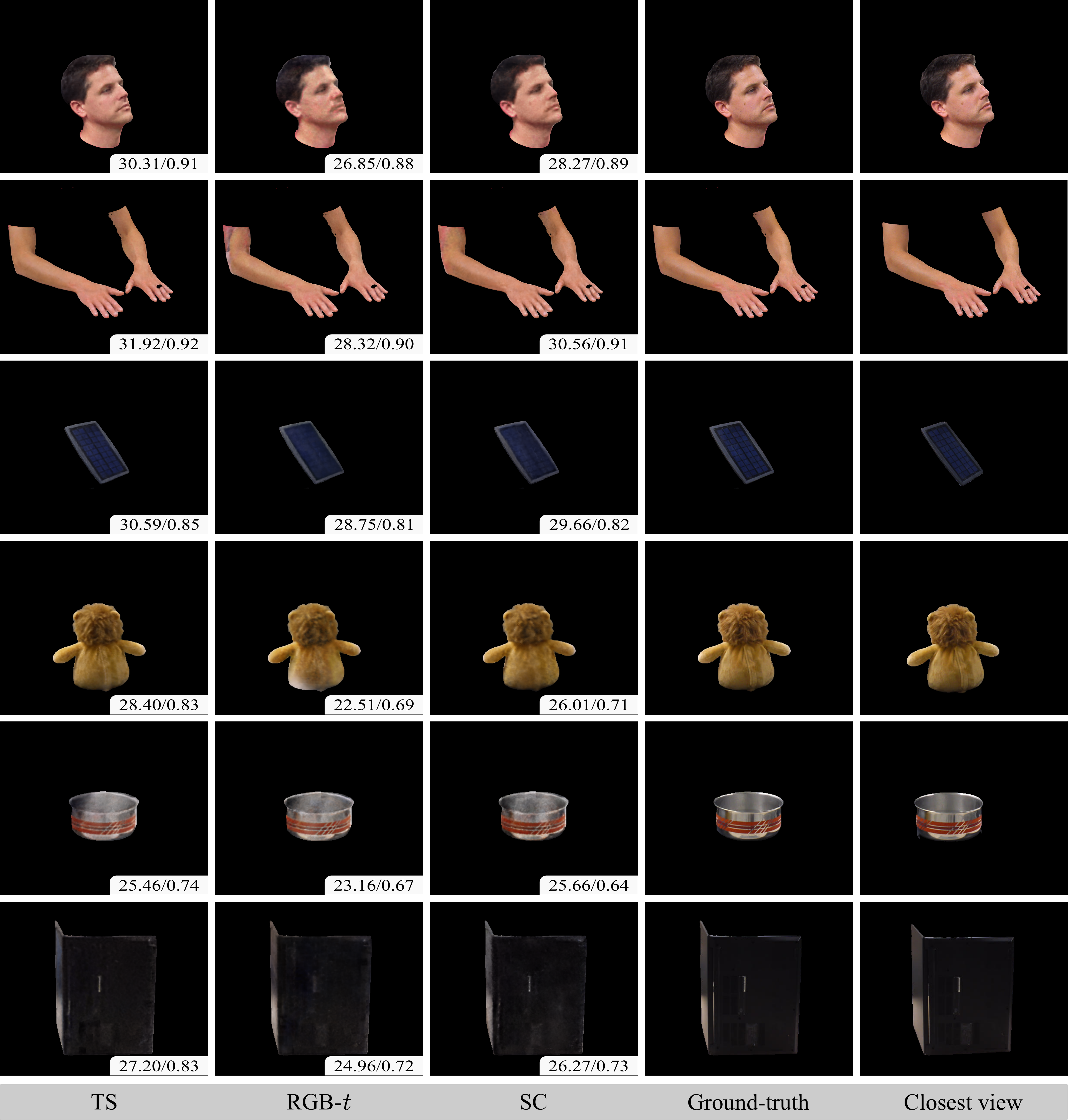}
    \caption{Reconstructions of a (left-out) RGB image from multi-modal neural scene representations trained on RGB and thermal data, arising from the four strategies that we compare. FT is left out since its RGB component is similar to TS. For each view, we also report PSNR and SSIM (higher is better). \textit{Closest view} denotes the nearest image in the training set.}
    \label{fig:results_rgb}
\end{figure}

\subsection{Evaluation on RGB Images}
\label{subsec:results_rgb}
Ideally, a multi-modal neural scene representation should be able to reconstruct images at least as good as its uni-modal counterpart.
In this section, we will assess the performance of the proposed strategies on RGB images.
Since FT's RGB component is similar to TS (they only differ in the number of iterations trained), we only show results for TS, RGB-$t$, and SC in this section.

Quantitative and qualitative results can be found in Table \ref{tab:uni_vs_multi} and Fig. \ref{fig:results_rgb}.
As can be observed, our baseline strategy, TS, \textit{slightly} outperforms all the other strategies in both, PSNR and SSIM.
Moreover, SC maintains superior reconstruction quality in terms of PSNR and SSIM compared to RGB-$t$.
This result can be attributed to SC's non-interference with RGB densities, whereas RGB-$t$ integrates thermal and RGB densities, causing a mixture of information.
Ultimately, when comparing both strategies to TS (which was solely trained on RGB images), we find that SC achieves similar reconstruction quality, whereas RGB-$t$ lags slightly behind.
Note that those findings also match the presented qualitative results, where only small differences between TS, RGB-$t$, and SC can be observed with the human eye.

\begin{table}[t]
 \caption{Quantitative results on (a) NIR and (b) RGB images, measured using PSNR and SSIM (higher is better). Results were obtained from NeRFs trained on RGB and NIR data. FT is left out since its RGB component is similar to TS.}
    \begin{subtable}{.54\linewidth}
      \centering
       \begin{adjustbox}{width=\linewidth}
        \begin{tabular}{lcc|cc|cc|cc}
            \toprule
             & \multicolumn{2}{c|}{TS} & \multicolumn{2}{c|}{FT} & \multicolumn{2}{c|}{RGB-NIR} & \multicolumn{2}{c}{SC} \\
             \cmidrule(lr){2-3} \cmidrule(lr){4-5} \cmidrule(lr){6-7} \cmidrule(lr){8-9}
             & PSNR $\uparrow$ & SSIM $\uparrow$ & PSNR $\uparrow$ & SSIM $\uparrow$ & PSNR $\uparrow$ & SSIM $\uparrow$ & PSNR $\uparrow$ & SSIM $\uparrow$ \\
            \midrule
            \textsc{Snail} & 36.55 & \textbf{0.97} & 32.32 & 0.95 & \textbf{36.70} & \textbf{0.97} & 35.55 & 0.96 \\
            \textsc{Bear} & \textbf{37.15} & \textbf{0.97} & 35.01 & 0.95 & 37.01 & 0.96 & 36.05 & 0.96 \\
            \textsc{Elephant} & \textbf{35.11} & \textbf{0.97} & 33.60 & 0.96 & 35.09 & \textbf{0.97} & 34.99 & 0.95 \\
            \midrule\midrule
            & \textbf{36.27} & \textbf{0.97} & 33.64 & 0.95 & \textbf{36.27} & \textbf{0.97} & 35.53 & 0.96 \\
            \bottomrule
        \end{tabular}
    \end{adjustbox}
    \caption{NIR reconstruction quality.}
    \end{subtable}\hfill
    \begin{subtable}{.42\linewidth}
      \centering
    \begin{adjustbox}{width=1.01\linewidth}
        \begin{tabular}{lcc|cc|cc}
            \toprule
             & \multicolumn{2}{c|}{TS} & \multicolumn{2}{c|}{RGB-NIR} & \multicolumn{2}{c}{SC} \\
             \cmidrule(lr){2-3} \cmidrule(lr){4-5} \cmidrule(lr){6-7}
             & PSNR $\uparrow$ & SSIM $\uparrow$ & PSNR $\uparrow$ & SSIM $\uparrow$ & PSNR $\uparrow$ & SSIM $\uparrow$ \\
            \midrule
            \textsc{Snail} & \textbf{39.94} & \textbf{0.97} & 37.31 & 0.96 & 38.03 & \textbf{0.97} \\
            \textsc{Bear} & \textbf{38.10} & \textbf{0.96} & 36.29 & \textbf{0.96} & 37.93 & \textbf{0.96} \\
            \textsc{Elephant} & \textbf{39.71} & \textbf{0.97} & 38.07 & 0.96 & 39.10 & \textbf{0.97} \\
            \midrule\midrule
             & \textbf{39.25} & \textbf{0.97} & 37.22 & 0.96 & 38.35 & \textbf{0.97} \\
            \bottomrule
        \end{tabular}
    \end{adjustbox}
    \caption{RGB reconstruction quality.}
    \end{subtable} 
    \label{tab:results_nir}
\end{table}

\begin{figure}[t]
    \centering
    \includegraphics[width=\textwidth]{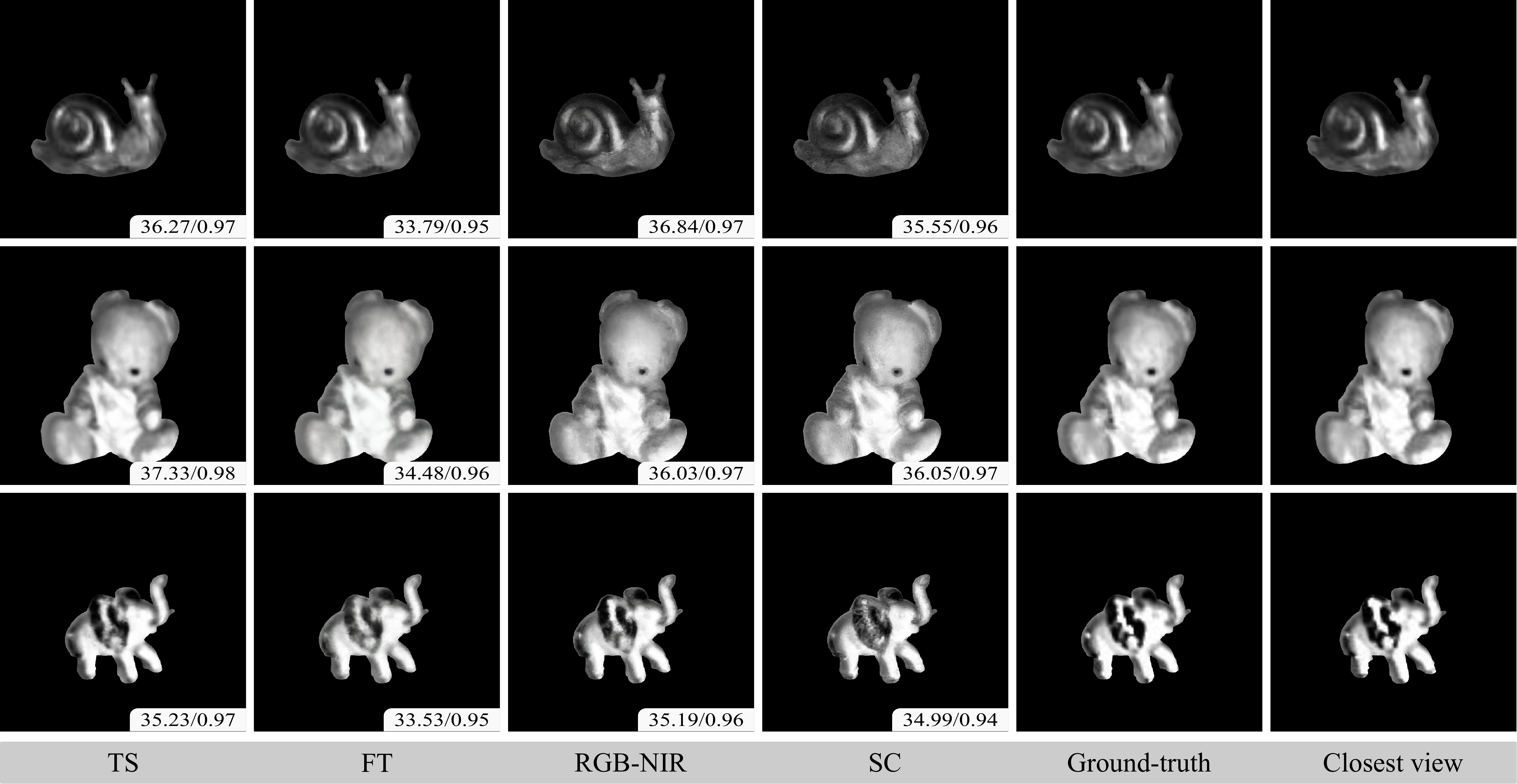}
    \caption{Reconstructions of a (left-out) NIR image from multi-modal neural scene representations trained on RGB and NIR data, arising from the four strategies that we compare. For each view, we also report PSNR and SSIM (higher is better). \textit{Closest view} denotes the nearest image in the training set.}
    \label{fig:results_rgb_nir_nir}
\end{figure}

 \begin{figure}[t]
    \centering
    \includegraphics[width=\textwidth]{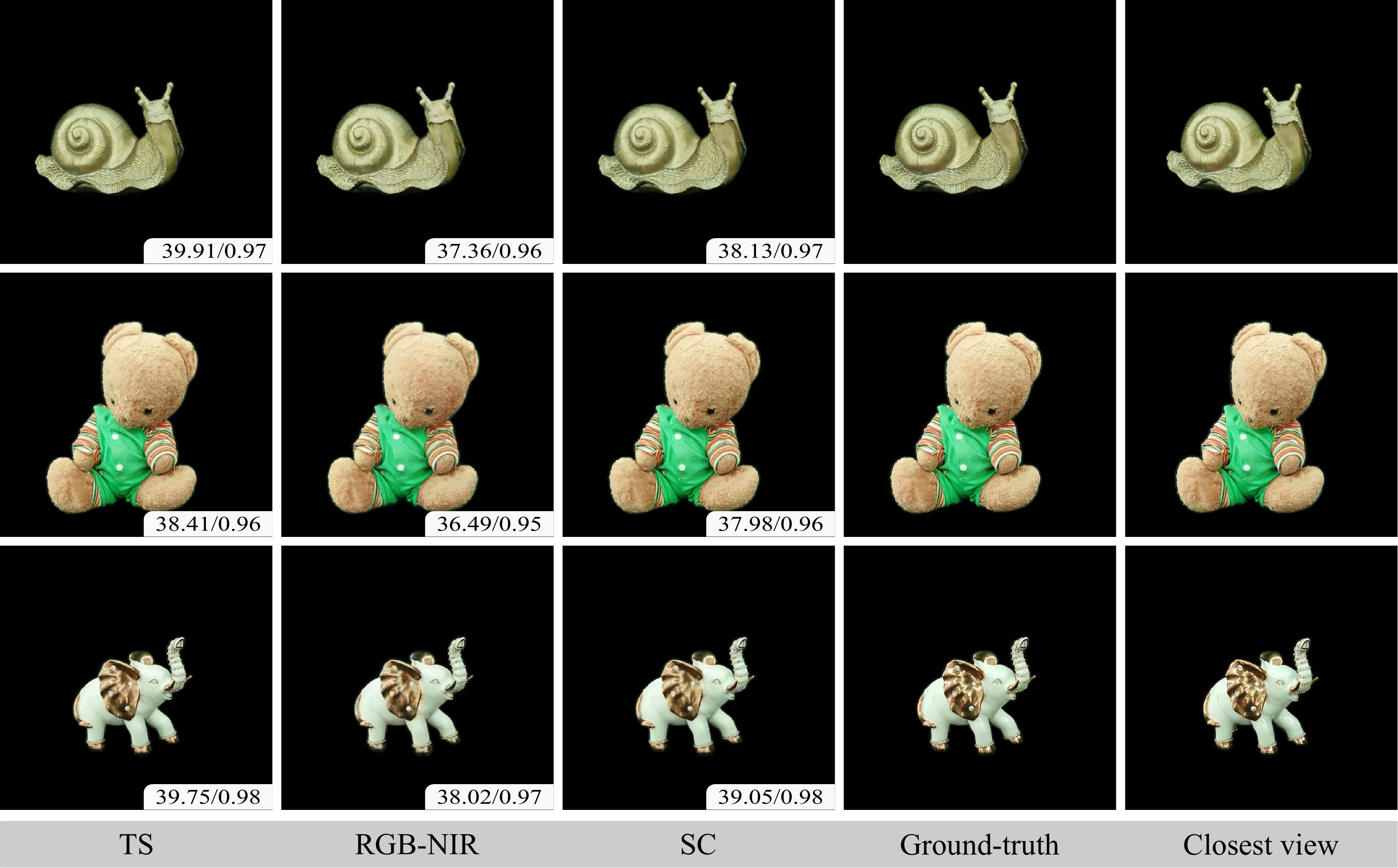}
    \caption{Reconstructions of a (left-out) RGB image from multi-modal neural scene representations trained on RGB and NIR data, arising from the four strategies that we compare. FT is left out since its RGB component is similar to TS. For each view, we also report PSNR and SSIM (higher is better). \textit{Closest view} denotes the nearest image in the training set.}
    \label{fig:results_rgb_nir_rgb}
\end{figure}

\subsection{Evaluation on Other Modalities}
\label{subsec:results_other}
Lastly, we also report results on multi-modal NeRFs learnt from RGB and near-infrared (NIR) images, and RGB and depth maps.
Due to space constraints, results for the latter can be found in the supp. material. 
However, they match exactly what we present here.

For all experiments, we used three randomly selected forward-facing scenes from the multi-sensor dataset proposed in \cite{voynov2023}.
Each scene contains 100 images, and we took images captured with the Huawei Mate 30 Pro (with a resolution of $7296\times 5472$ for RGB and $240\times 180$ for NIR images). 
Note that, due to the different resolutions, a direct comparison between results obtained from RGB and thermal (see Sections \ref{subsec:results_thermal} and \ref{subsec:results_rgb}) and RGB and NIR would be unfair.
However, since we are anyway only interested in a relative comparison between proposed strategies, having different resolutions is not a problem.

\textbf{Evaluation on NIR images.}
We report results in Table \ref{tab:results_nir}(a) and Fig. \ref{fig:results_rgb_nir_nir}.
Considering NIR reconstruction quality, RGB-X (denoted as RGB-NIR in the following) performs best on average, but, as opposed to RGB and thermal, this time on par with TS.
This is understandable given the fact that NIR images do not have as static and texture-less background as thermal images, effectively mitigating the problems described in Section \ref{subsec:results_thermal}.
All in all, we observe that incorporating NIR images into a multi-modal NeRF seems to be easier (\ie, less dependent on the employed strategy), which (i) confirms our hypothesis that modeling thermal images seems to be hard, and (ii) justifies the usage of thermal imagery as a challenging benchmark for evaluating the proposed strategies. 

\textbf{Evaluation on RGB images.}
Quantitative and qualitative results on RGB reconstruction quality can be found in Table \ref{tab:results_nir}(b) and Fig. \ref{fig:results_rgb_nir_rgb}.
We observe a very similar trend as for multi-modal NeRFs learnt from RGB and thermal images: TS performs best in terms of both, PSNR and SSIM, followed by SC and RGB-NIR.

\section{Discussion and Limitations}
Based on our analysis, RGB-X consistently outperforms all strategies in reconstructing thermal images, and performs on par with our baseline, TS, when it comes to NIR images and depth maps.
Taking into account RGB reconstruction quality, SC slightly surpasses RGB-X when trained on RGB and thermal, NIR, or depth data.
Moreover, our analysis reveals that, especially for low-texture images such as thermal images, allowing NeRF's volume densities to be influenced by the second modality (as in RGB-X) helps a lot in learning multi-modal neural scene representations, effectively causing a mixture of information.
All in all, due to the fact that RGB-X still yields compelling results on RGB data (although it can not meet the quality of its uni-modal counterpart), we conclude that RGB-X seems to be well suited for multi-modal neural scene representations.

\textbf{Limitations.}
We also want to note a limitation of the present study, predominantly from a practical point of view.
First off, although offline cross-modality calibration provides almost perfect alignments between RGB and thermal images, it prevents us from building NeRFs of uncalibrated, in-the-wild images.
This could be approached by integrating learning-based calibration methods as proposed in \cite{Poggi2022,Zhu2023}.
However, since the focus of this work is clearly \textit{not} on (learning) cross-modality calibration but rather on the systematic evaluation of different strategies to integrate multi-modality into NeRFs, we leave this for future work.

\section{Conclusion and Future Work}
In this paper, we have systematically compared four different strategies of how to include image data from different modalities, other than RGB, into a single scene representation, ultimately aiming for multi-modal neural scene representations.
Based on RGB data and a NeRF-like scene representation as our base model, we propose to include images from a second modality using (1) training from scratch (TS), (2) fine-tuning (FT), (3) adding a second branch (RGB-X), and (4) adding a separate component (SC) to the base model.
The analysis of the four strategies is based on a newly captured, object-centric dataset, named \textit{ThermalMix}, which consists of 360 multi-view RGB and thermal images.
The dataset includes six common objects in total, three of them captured as forward-facing scenes and three as 360-degree scenes, and it is the first to provide near-perfect alignments between RGB and thermal images.
\textit{ThermalMix} is publicly available, and we believe it serves as a challenging benchmark not only for reconstruction tasks, but also for learnable cross-modality calibration.

Our findings indicate that RGB-X stands out for its thermal reconstruction capabilities while also delivering compelling RGB reconstructions.
Finally, we could also show that our results generalize to other modalities, including NIR images and depth maps, leading to the conclusion that RGB-X indeed seems to be well-suited for building general multi-modal neural scene representations.

\textbf{Future work.}
Our future work is primarily concerned with the incorporation and evaluation of learning-based schemes for online cross-modality calibration.

\section*{Acknowledgements}
We would like to thank Ian Marius Peters, Bernd Doll, and Oleksandr Mashkov for valuable discussions and access to the thermal camera.
This work was funded by the German Federal Ministry of Education and Research (BMBF), FKZ: 01IS22082 (IRRW). The authors are responsible for the content of this publication.

\bibliographystyle{splncs04}
\bibliography{main}

\begin{thebibliography}{10}
\providecommand{\url}[1]{\texttt{#1}}
\providecommand{\urlprefix}{URL }
\providecommand{\doi}[1]{https://doi.org/#1}

\bibitem{aggarwal2023}
Aggarwal, A.K.: Thermal imaging for cancer detection. Imaging Radiat Res
  \textbf{6} (2023)

\bibitem{akula2011}
Akula, A., Ghosh, R., Sardana, H.K.: Thermal imaging and its application in
  defence systems. In: AIP Conf Proc (2011)

\bibitem{alldieck2016}
Alldieck, T., Bahnsen, C.H., Moeslund, T.B.: Context-aware fusion of rgb and
  thermal imagery for traffic monitoring. Sensors  \textbf{16} (2016)

\bibitem{Koc2021}
Ceyhun, K., Ozgun, P., Sultan, T.: 3d mesh model generation from ct and mri
  data. In: IEEE BigData 2021 (2021)

\bibitem{Deng2021}
Deng, K., Liu, A., Zhu, J.Y., Ramanan, D.: Depth-supervised nerf: Fewer views
  and faster training for free. In: CVPR (2021)

\bibitem{Meng2023}
Dongdong, M., Sheng, L., Bin, S., Hao, W., Suqing, T., Wenjun, M., Guoping, W.,
  Xueqing, Y.: 3d reconstruction-oriented fully automatic multi-modal tumor
  segmentation by dual attention-guided vnet. Vis Comput  \textbf{39} (2023)

\bibitem{gade2014}
Gade, R., Moeslund, T.B.: Thermal cameras and applications: A survey. Mach Vis
  Appl  (2014)

\bibitem{messina2020}
Gaetano, M., Giuseppe, M.: Applications of uav thermal imagery in precision
  agriculture: State of the art and future research outlook. Remote Sens
  \textbf{12} (2020)

\bibitem{gao2023}
Gao, K., Gao, Y., He, H., Lu, D., Xu, L., Li, J.: Nerf: Neural radiance field
  in 3d vision, a comprehensive review. arXiv:2210.00379  (2023)

\bibitem{gowen2010}
Gowen, A.A., Tiwari, B.K., Cullen, P.J., McDonnell, K., O'Donnell, C.P.:
  Applications of thermal imaging in food quality and safety assessment. Trends
  Food Sci  \textbf{21} (2010)

\bibitem{Zhu2023}
Haidong, Z., Yuyin, S., Chi, L., Lu, X., Jiajia, L., Nan, Q., Ramkant, N.,
  Cheng-Hao, K.: Multimodal neural radiance field. In: ICRA (2023)

\bibitem{Han2019}
Han, W., Liu, X., Song, S., Meng, M.Q.H.: 3d reconstruction of dense model
  based on the sparse frames using rgbd camera. In: ROBIO (2019)

\bibitem{Zhu2022}
Haoyi, Z.: X-nerf: Explicit neural radiance field for multi-scene 360°
  insufficient rgb-d views. In: WACV (2023)

\bibitem{hassan2024}
Hassan, M., Forest, F., Fink, O., Mielle, M.: Thermonerf: Multimodal neural
  radiance fields for thermal novel view synthesis. arXiv:2403.12154  (2024)

\bibitem{Hedman2021}
Hedman, P., Srinivasan, P.P., Mildenhall, B., Barron, J.T., Debevec, P.: Baking
  neural radiance fields for real-time view synthesis. In: ICCV (2021)

\bibitem{Herau2023}
Herau, Q., Piasco, N., Bennehar, M., Roldão, L., Tsishkou, D., Migniot, C.,
  Vasseur, P., Demonceaux, C.: Moisst: Multi-modal optimization of implicit
  scene for spatiotemporal calibration. In: IROS (2023)

\bibitem{huo2023}
Huo, D., Wang, J., Qian, Y., Yang, Y.H.: Glass segmentation with rgb-thermal
  image pairs. IEEE Trans Image Process  \textbf{32} (2023)

\bibitem{ishimwe2014}
Ishimwe, R., Abutaleb, K., Ahmed, F.: Applications of thermal imaging in
  agriculture--a review. ARS  \textbf{3} (2014)

\bibitem{jones2004}
Jones, H.G.: Application of thermal imaging and infrared sensing in plant
  physiology and ecophysiology. Adv Bot Res  \textbf{41} (2004)

\bibitem{Joshi2016}
Joshi, N.P., Baumann, M., Ehammer, A., Fensholt, R., Grogan, K., Hostert, P.,
  Rudbeck, M.J., Kuemmerle, T., Meyfroidt, P., Mitchard, E.T.A., Reiche, J.,
  Ryan, C.M., Waske, B.: A review of the application of optical and radar
  remote sensing data fusion to land use mapping and monitoring. Remote Sens
  \textbf{8} (2016)

\bibitem{khanal2017}
Khanal, S., Fulton, J.P., Shearer, S.A.: An overview of current and potential
  applications of thermal remote sensing in precision agriculture. Comput
  Electron Agric  \textbf{139} (2017)

\bibitem{Kingma2014}
Kingma, D.P., Ba, J.: Adam: A method for stochastic optimization. In: ICLR
  (2015)

\bibitem{li2023}
Li, G., Lin, Y., Ouyang, D., Li, S., Luo, X., Qu, X., Pi, D., Li, S.E.: A
  rgb-thermal image segmentation method based on parameter sharing and
  attention fusion for safe autonomous driving. IEEE Trans Intell Transp Syst
  (2023)

\bibitem{lin2024}
Lin, Y.Y., Pan, X.Y., Fridovich-Keil, S., Wetzstein, G.: Thermalnerf: Thermal
  radiance fields. In: ICCP (2024)

\bibitem{Xliu2023}
Liu, X., Li, Y., Teng, Y., Bao, H., Zhang, G., Zhang, Y., Cui, Z.: Multi-modal
  neural radiance field for monocular dense slam with a light-weight tof
  sensor. In: ICCV (2023)

\bibitem{Mildenhall2020}
Mildenhall, B., Srinivasan, P.P., Tancik, M., Barron, J.T., Ramamoorthi, R.,
  Ng, R.: Nerf: Representing scenes as neural radiance fields for view
  synthesis. In: ECCV (2020)

\bibitem{Mueller2022}
Müller, T., Evans, A., Schied, C., Keller, A.: Instant neural graphics
  primitives with a multiresolution hash encoding. ACM Trans Graph  \textbf{41}
  (2022)

\bibitem{Petneházy2023}
Örs Petneházy, Rück, S., Sós, E., Reinitz, L.Z.: 3d reconstruction of the
  blood supply in an elephant’s forefoot using fused ct and mri sequences.
  Animals  \textbf{13} (2023)

\bibitem{pineda2020}
Pineda, M., Barón, M., Pérez-Bueno, M.L.: Thermal imaging for plant stress
  detection and phenotyping. Remote Sens  \textbf{13} (2021)

\bibitem{Poggi2022}
Poggi, M., Ramirez, P.Z., Tosi, F., Salti, S., Mattoccia, S., Stefano, L.D.:
  Cross-spectral neural radiance fields. In: I3DV (2022)

\bibitem{rai2017}
Rai, M.K., Maity, T., Yadav, R.K.: Thermal imaging system and its real time
  applications: a survey. J Eng Technol  \textbf{25} (2017)

\bibitem{raju2023}
Raju, V.B., Imtiaz, M.H., Sazonov, E.: Food image segmentation using
  multi-modal imaging sensors with color and thermal data. Sensors  \textbf{23}
  (2023)

\bibitem{ring2012}
Ring, E.F.J., Ammer, K.: Infrared thermal imaging in medicine. Physiol Meas
  \textbf{33} (2012)

\bibitem{Schoenberger2016}
Sch\"{o}nberger, J.L., Frahm, J.M.: Structure-from-motion revisited. In: CVPR
  (2016)

\bibitem{shaikh2019}
Shaikh, S., Akhter, N., Manza, R.: Current trends in the application of thermal
  imaging in medical condition analysis. IJITEE  \textbf{8} (2019)

\bibitem{shivakumar2020}
Shivakumar, S.S., Rodrigues, N., Zhou, A., Miller, I.D., Kumar, V., Taylor,
  C.J.: Pst900: Rgb-thermal calibration, dataset and segmentation network. In:
  ICRA (2020)

\bibitem{Shum2000}
Shum, H.Y., Kang, S.B.: Review of image-based rendering techniques. In: VCIP
  (2000)

\bibitem{deSouza2023}
de~Souza, M.A., Cordeiro, D.C.A., de~Oliveira, J., de~Oliveira, M.F.A.,
  Bonafini, B.L.: 3d multi-modality medical imaging: Combining anatomical and
  infrared thermal images for 3d reconstruction. Sensors  \textbf{23} (2023)

\bibitem{still2019}
Still, C.J., Powell, R.L., Aubrecht, D.M., Kim, Y., Helliker, B.R., Roberts,
  D.A., Richardson, A.D., Goulden, M.L.: Thermal imaging in plant and ecosystem
  ecology: applications and challenges. Ecosphere  \textbf{10} (2019)

\bibitem{swamidoss2021}
Swamidoss, I.N., Amro, A.B., Sayadi, S.: Systematic approach for thermal
  imaging camera calibration for machine vision applications. Optik
  \textbf{247} (2021)

\bibitem{Tancik2020}
Tancik, M., Srinivasan, P.P., Mildenhall, B., Fridovich-Keil, S., Raghavan, N.,
  Singhal, U., Ramamoorthi, R., Barron, J.T., Ng, R.: Fourier features let
  networks learn high frequency functions in low dimensional domains. In:
  NeurIPS (2020)

\bibitem{tao2024}
Tang, T., Wang, G., Lao, Y., Chen, P., Liu, J., Lin, L., Yu, K., Liang, X.:
  Alignmif: Geometry-aligned multimodal implicit field for lidar-camera joint
  synthesis. arXiv:2402.17483  (2024)

\bibitem{MoosaviTayebi2015}
Tayebi, R.M., Wirza, R., Sulaiman, P.S., Dimon, M.Z., Khalid, F., Al-Surmi,
  A.A., Mazaheri, S.: 3d multimodal cardiac data reconstruction using
  angiography and computerized tomographic angiography registration. J
  Cardiothorac Surg  \textbf{10} (2015)

\bibitem{vadivambal2011}
Vadivambal, R., Jayas, D.S.: Applications of thermal imaging in agriculture and
  food industry -- a review. Food Bioproc Tech  \textbf{4} (2011)

\bibitem{voynov2023}
Voynov, O., Bobrovskikh, G., Karpyshev, P., Galochkin, S., Ardelean, A.T.,
  Bozhenko, A., Karmanova, E., Kopanev, P., Labutin-Rymsho, Y., Rakhimov, R.,
  Safin, A., Serpiva, V., Artemov, A., Burnaev, E., Tsetserukou, D., Zorin, D.:
  Multi-sensor large-scale dataset for multi-view 3d reconstruction. In: CVPR
  (2023)

\bibitem{Wakeford2019}
Wakeford, Z.E., Chmielewska, M., Hole, M.J., Howell, J.A., Jerram, D.A.:
  Combining thermal imaging with photogrammetry of an active volcano using uav:
  an example from stromboli, italy. The Photogrammetric Record  \textbf{34}
  (2019)

\bibitem{wang2023}
Wang, D., Zhang, T., Abboud, A., Süsstrunk, S.: Inpaintnerf360: Text-guided 3d
  inpainting on unbounded neural radiance fields. arXiv:2305.15094  (2023)

\bibitem{wilson2023}
Wilson, A.N., Gupta, K., Koduru, B.H., Kumar, A., Jha, A., Cenkeramaddi, L.R.:
  Recent advances in thermal imaging and its applications using machine
  learning: A review. IEEE Sens J  \textbf{23} (2023)

\bibitem{xu2024}
Xu, J., Liao, M., Prabhakar, K.R., Patel, V.M.: Leveraging thermal modality to
  enhance reconstruction in low-light conditions. In: ECCV (2024)

\bibitem{yang2023}
Yang, X., Guo, R., Li, H.: Comparison of multimodal rgb-thermal fusion
  techniques for exterior wall multi-defect detection. JIIR  \textbf{2} (2023)

\bibitem{Furukawa2015}
Yasutaka, F., Carlos, H.: Multi-view stereo: A tutorial. Foundations and Trends
  in Computer Graphics and Vision  \textbf{9} (2015)

\bibitem{ye2024}
Ye, T., Wu, Q., Deng, J., Liu, G., Liu, L., Xia, S., Pang, L., Yu, W., Pei, L.:
  Thermal-nerf: Neural radiance fields from an infrared camera.
  arXiv:2403.10340  (2024)

\bibitem{Yu2022}
Yu, A., Fridovich-Keil, S., Tancik, M., Chen, Q., Recht, B., Kanazawa, A.:
  Plenoxels: Radiance fields without neural networks. In: CVPR (2022)

\bibitem{zhang2023}
Zhang, Q., Wang, B.H., Yang, M.C., Zou, H.: Mmnerf: Multi-modal and multi-view
  optimized cross-scene neural radiance fields. IEEE Access  \textbf{11} (2023)

\bibitem{Zollhöfer2018}
Zollhöfer, M., Stotko, P., Gorlitz, A., Theobalt, C., Nießner, M., Klein, R.,
  Kolb, A.: State of the art on 3d reconstruction with rgb-d cameras. Computer
  Graphics Forum  \textbf{37} (2018)

\bibitem{stumper2015}
Štumper, M., Kraus, J.: Thermal imaging in aviation. MAD  \textbf{3} (2015)

\end{thebibliography}

\clearpage

\appendix

\title{Exploring Multi-modal Neural Scene Representations With Applications \\ on Thermal Imaging \\ \vspace{0.3cm} \large{--- Supplementary Material ---}}

\titlerunning{Exploring Multi-modal Neural Scene Representations}

\author{Mert Özer\thanks{Authors contributed equally to this work.} \and
Maximilian Weiherer\printfnsymbol{1} \and
Martin Hundhausen \and
Bernhard Egger}

\authorrunning{M.~Özer et al.}

\institute{Friedrich-Alexander-Universität Erlangen-Nürnberg\\
\email{firstname.lastname@fau.de}}

\maketitle

\noindent In this supplementary material, we (i) demonstrate that it is challenging to compute (reliable) camera poses from thermal images (Section \ref{sec:camera_pose}), (ii) investigate the difference in geometry between RGB and thermal images (Section \ref{sec:geometry}), (iii) provide an ablation on the weights $w_c$ and $w_t$ in Eq. (\ref{eq:combined_loss}) of the main paper (Section \ref{sec:ablation}), (iv) give more details on how we align RGB and thermal images during cross-modality calibration (Section \ref{sec:calibration}), (v) further analyze why 360-degree scenes are harder to optimize than forward-facing scenes (Section \ref{sec:challenges_360}), and (vi) present results on multi-modal NeRFs learned from RGB and depth maps (Section \ref{sec:results_depth}).

\begin{figure}
     \centering
     \includegraphics[width=0.65\linewidth]{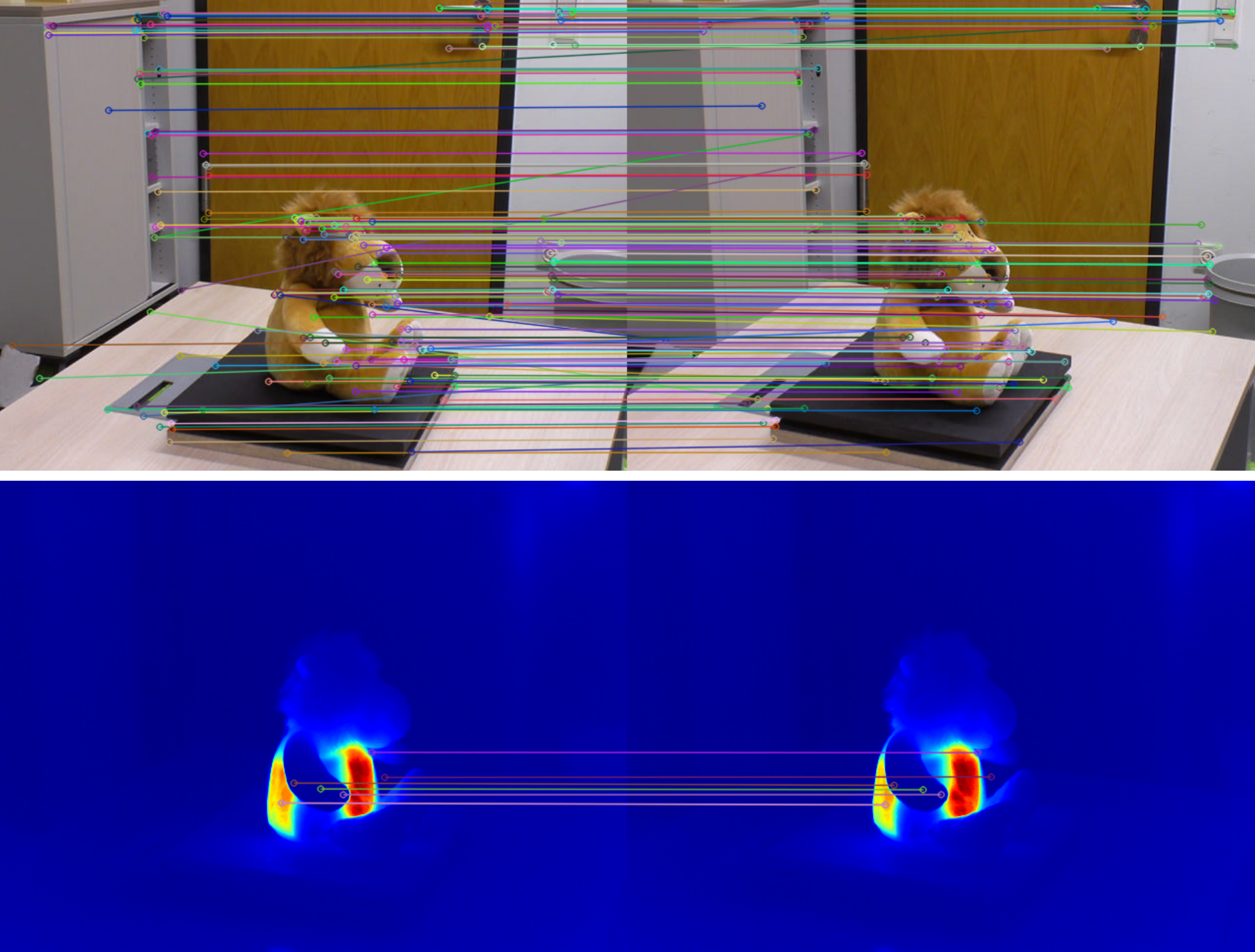}
     \caption{Demonstration of how challenging it is to compute \textit{reliable} camera poses from thermal images. We visualize feature correspondences between two views on RGB (top row) and thermal images (bottom row; found and matched using COLMAP \cite{Schoenberger2016}).}
     \label{fig:ft_rgb_thermal}
\end{figure}

\section{Camera Poses From Thermal Images}
\label{sec:camera_pose}
In Fig. \ref{fig:ft_rgb_thermal} we show matching features between two RGB images and corresponding thermal images found using COLMAP \cite{Schoenberger2016}.
For RGB, 127 reliable matches have been found, while only seven matches could be found on thermal images, rendering it almost impossible to compute a meaningful camera pose.
This demonstrates that classical feature extractors (and hence, structure-from-motion techniques) struggle with thermal images, ultimately requiring new, specialized methods for camera pose estimation from non-RGB images.

\begin{figure}[t]
    \includegraphics[width=\linewidth]{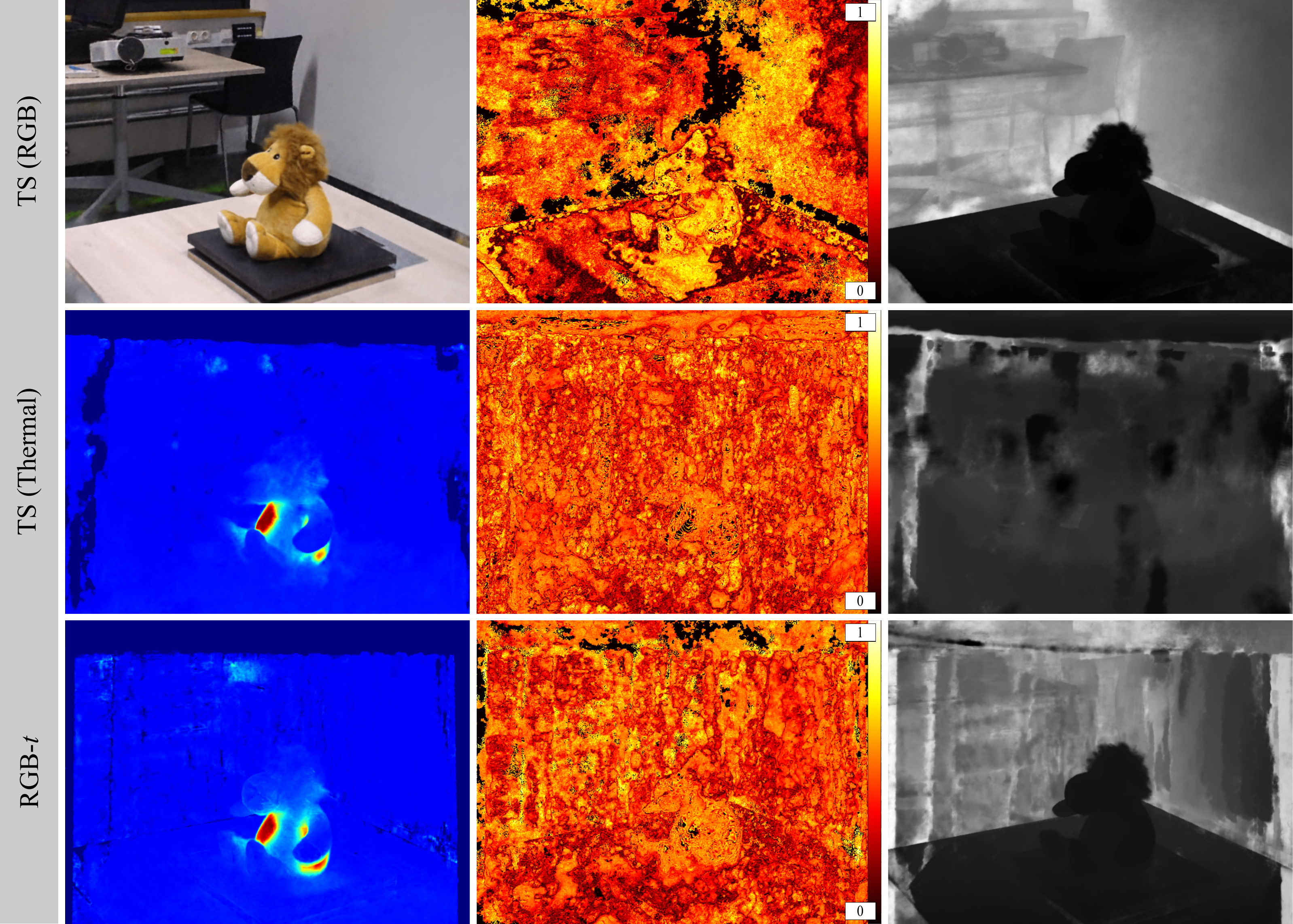}
    \caption{Comparison of the reconstructed geometry in RGB (first row) and thermal images (second and third row), shown on TS and RGB-$t$. The first column shows novel renderings, the second column visualizes accumulated densities for each pixel along its respective ray, and the third column depicts estimated depth maps. As can be observed clearly, thermal-derived geometry greatly benefits from utilizing RGB densities (especially seen in the depth maps arising from TS trained solely on thermal images (second row) and depth maps produced by RGB-$t$ (third row), which, contrary to TS, incorporates RGB information).}
    \label{fig:one}
\end{figure}

\begin{figure}
    \centering
    \includegraphics[width=0.65\linewidth]{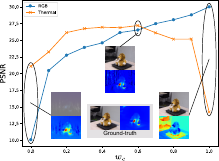}
    \caption{Ablation on the weights $w_c$ and $w_t$ of Eq. (\ref{eq:combined_loss}) in the main paper. We evaluate for $w_c\in\{0, 0.1, 0.2, \dots, 1\}$ and set $w_t=1-w_c$. Models were trained with RGB-$t$.}
    \label{fig:weights_ablation}
\end{figure}

\section{Different Geometry From RGB and Thermal Images}
\label{sec:geometry}
As noted in Section \ref{sec:method} of the main paper and based on our analysis, we observe significant differences in scene densities when comparing RGB and thermal modalities, leading to variations in the reconstructed geometry of the scene.
As illustrated in Fig. \ref{fig:one}, the second column presents the accumulated densities for each pixel along its respective ray. This visualization highlights a distinct shift when transitioning from training exclusively with RGB data to training with thermal data from scratch (as done in TS). Notably, RGB-$t$ appears to strike a balance in these density distributions. It is evident that thermal images can benefit from utilizing RGB densities, leading to improved reconstruction quality.

Further insights are offered in the third column showing predicted depth maps.
As seen, depth maps elucidate that in thermal imaging, the uniform background tends to create the illusion of a single continuous surface across most of the image, except for the high-temperature regions on the object of interest.
This effect underscores the unique challenges and considerations when interpreting thermal imagery in contrast to RGB and the advantage of integrating RGB information for more effective reconstruction of thermal data.

\section{Ablation on Weights of Loss}
\label{sec:ablation}
We provide an ablation on $w_c$ and $w_t$ as used in Eq. (\ref{eq:combined_loss}) of the main paper in Fig. \ref{fig:weights_ablation}. 
Results were obtained using RGB-$t$ trained on the 360-degree \textsc{Lion} scene.
As can be observed, values of the two weights matter, but are not very sensitive. 

\begin{figure}
    \centering
    \includegraphics[width=\linewidth]{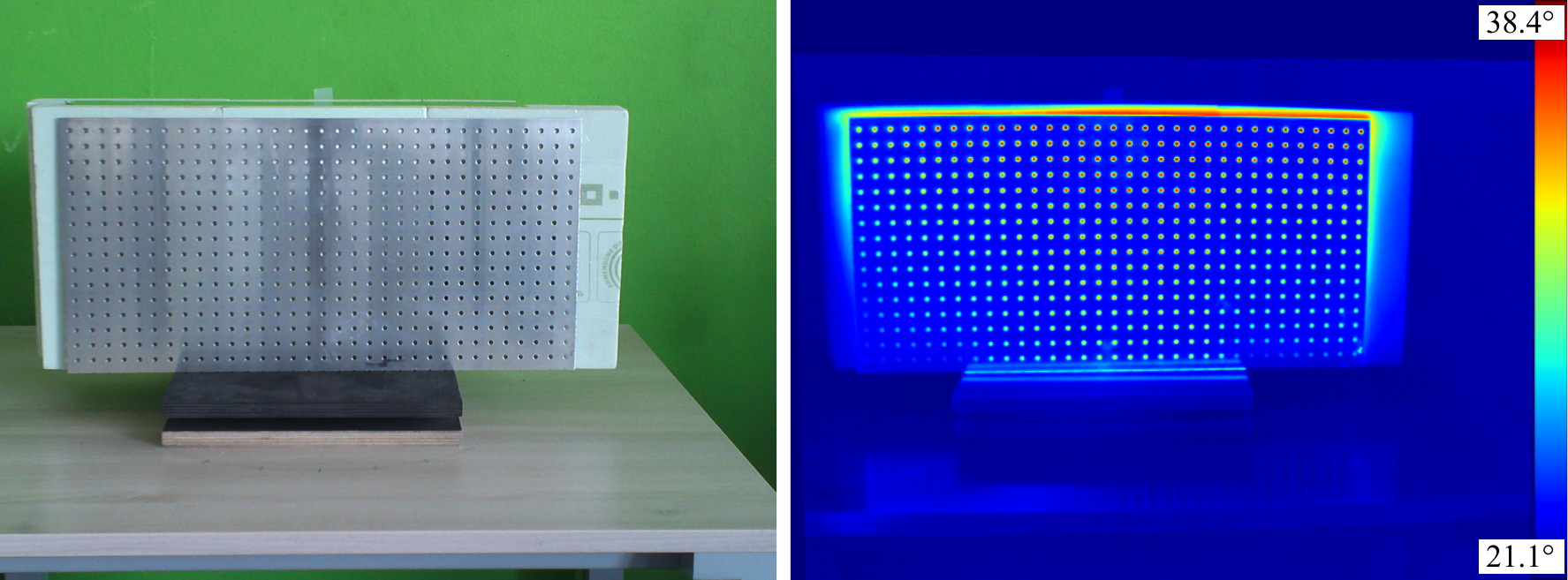}
     \caption{Calibration object used to align RGB and thermal images. We utilized a perforated, slightly warmed-up aluminum plate, which is visible in both, RGB and thermal images. To compute relative poses between RGB and infrared sensors, we identify the holes' midpoints as matching features across the two modalities.}
     \label{fig:calibration_object}
\end{figure}
 
\section{Details on Cross-Modality Calibration}
\label{sec:calibration}
As explained in Section \ref{sec:dataset} of the main paper, we align RGB and thermal images using a perforated aluminum plate (which we just bought in a local hardware store) as calibration object visible in both imaging modalities, see Fig. \ref{fig:calibration_object}.

We establish at least four point correspondences (as stated, we detect midpoints of the holes) between RGB and thermal images of the calibration object.
For corresponding points \((x_i, y_i)\) in the RGB image and \((x'_i, y'_i)\) in the thermal image, we calculate the homography matrix \( H \) using the following steps.

1. \textbf{Formulating equations:} Each set of corresponding points yields two equations:
   \[
   x'_i = \frac{h_{11} x_i + h_{12} y_i + h_{13}}{h_{31} x_i + h_{32} y_i + h_{33}}, \quad
   y'_i = \frac{h_{21} x_i + h_{22} y_i + h_{23}}{h_{31} x_i + h_{32} y_i + h_{33}}.
   \]

2. \textbf{Rearranging into linear system:} These equations are rearranged to a linear form \( Ax = b \), leading to the matrix equation:
   \[
   \begin{bmatrix}
   x_i & y_i & 1 & 0 & 0 & 0 & -x'_i x_i & -x'_i y_i \\
   0 & 0 & 0 & x_i & y_i & 1 & -y'_i x_i & -y'_i y_i 
   \end{bmatrix}
   \begin{bmatrix}
   h_{11} \\
   h_{12} \\
   h_{13} \\
   h_{21} \\
   h_{22} \\
   h_{23} \\
   h_{31} \\
   h_{32}
   \end{bmatrix}
   = 
   \begin{bmatrix}
   x'_i \\
   y'_i
   \end{bmatrix}.
   \]

3. \textbf{Solving with SVD:} The system is typically over-determined, so we apply Singular Value Decomposition (SVD) to find the solution. The homography matrix \( H \) is obtained from the last column of \( V \) (from the decomposition \( A = U \Sigma V^T \)) corresponding to the smallest singular value.
The resulting matrix \( H \) enables the transformation of coordinates from the RGB image plane to the thermal image plane, facilitating the accurate mapping of thermal images onto the RGB camera space.

\begin{figure}
    \centering
    \includegraphics[width=\linewidth]{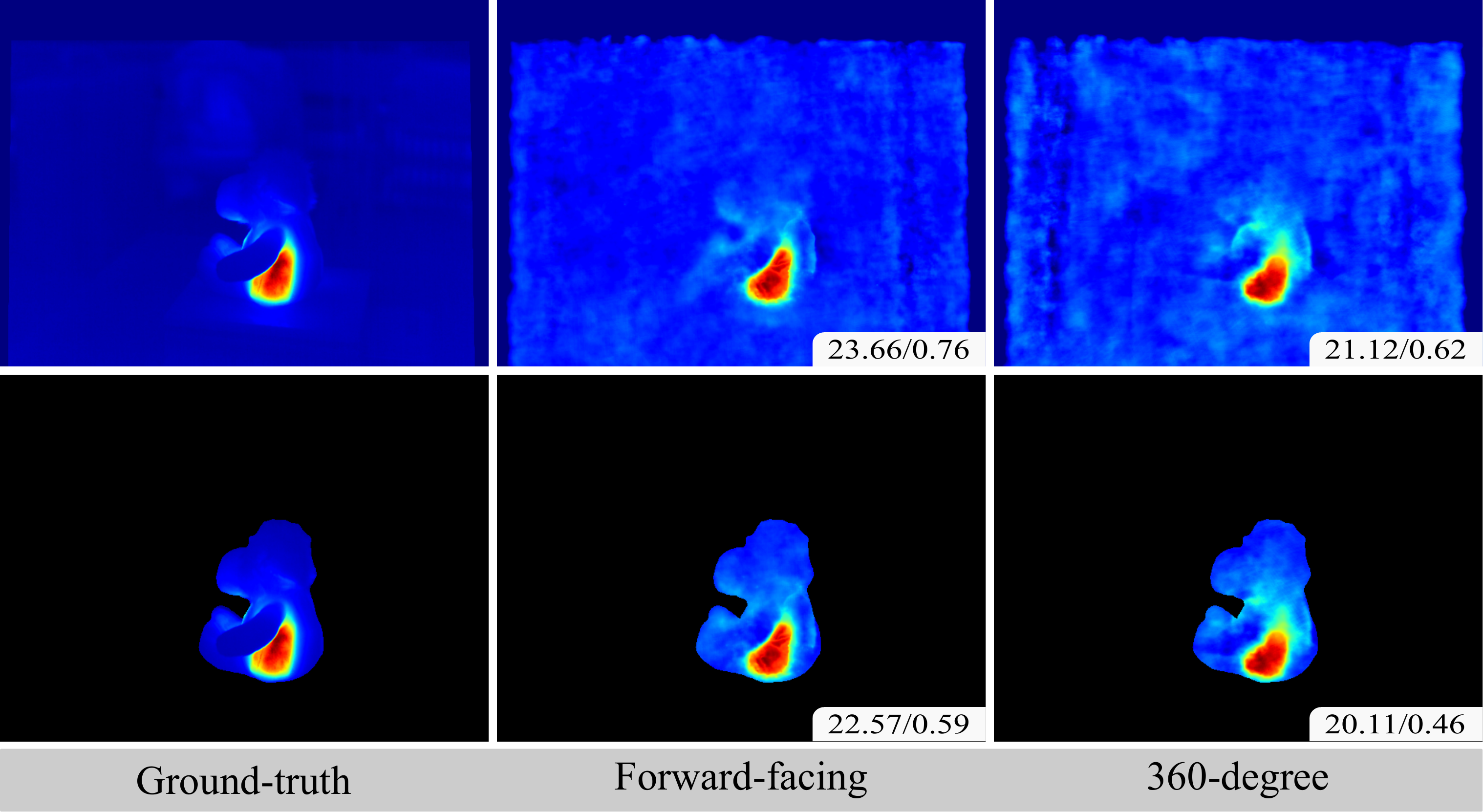}
    \caption{Novel views (second and third column) produced by TS trained on \textsc{Lion}. For the second column, we trained TS only on the forward-facing views from the dataset, while for the third column, we used all available images. Second row shows segmented predictions from the first row. We also report PSNR and SSIM (higher is better).}
    \label{fig:forward_vs_360}
\end{figure}

\section{Challenges on 360-Degree Scenes}
\label{sec:challenges_360}
Training NeRF on 360-degree scenes using thermal images presents unique complexities as noted in Section \ref{subsec:results_thermal} of the main paper. One primary challenge is the static appearance of thermal backgrounds, particularly in object-centered scenarios. In such cases, the background often appears uniform, creating an illusion that the object is rotating rather than the camera moving around the object. This phenomenon can mislead the network during training, allowing it to minimize loss by simply adjusting the background color. In an experiment, where TS was trained on half and the entirety of a 360-degree \textsc{Lion} scene for 10,000 iterations (as shown in Fig. \ref{fig:forward_vs_360}), we observed that forward-facing training significantly reduces the occurrence of floaters present in the scene. One plausible reason why forward-facing scenes may be more effective than 360-degree scenes for depth estimation is the nature of background representation. In 360-degree scenes, the background often appears as a static surface, very similar at both the front and back of the central object, akin to a curtain enveloping the object (see Fig. \ref{fig:forward_vs_360}). This uniformity can obscure the object, making depth perception challenging. In contrast, forward-facing scenes typically exhibit this static background effect only behind the object, not in front, allowing for clearer distinction and depth estimation of the central subject. Once again, this idea leads to the importance of RGB information utilization for thermal reconstruction.

\begin{figure}
    \centering
    \includegraphics[width=\linewidth]{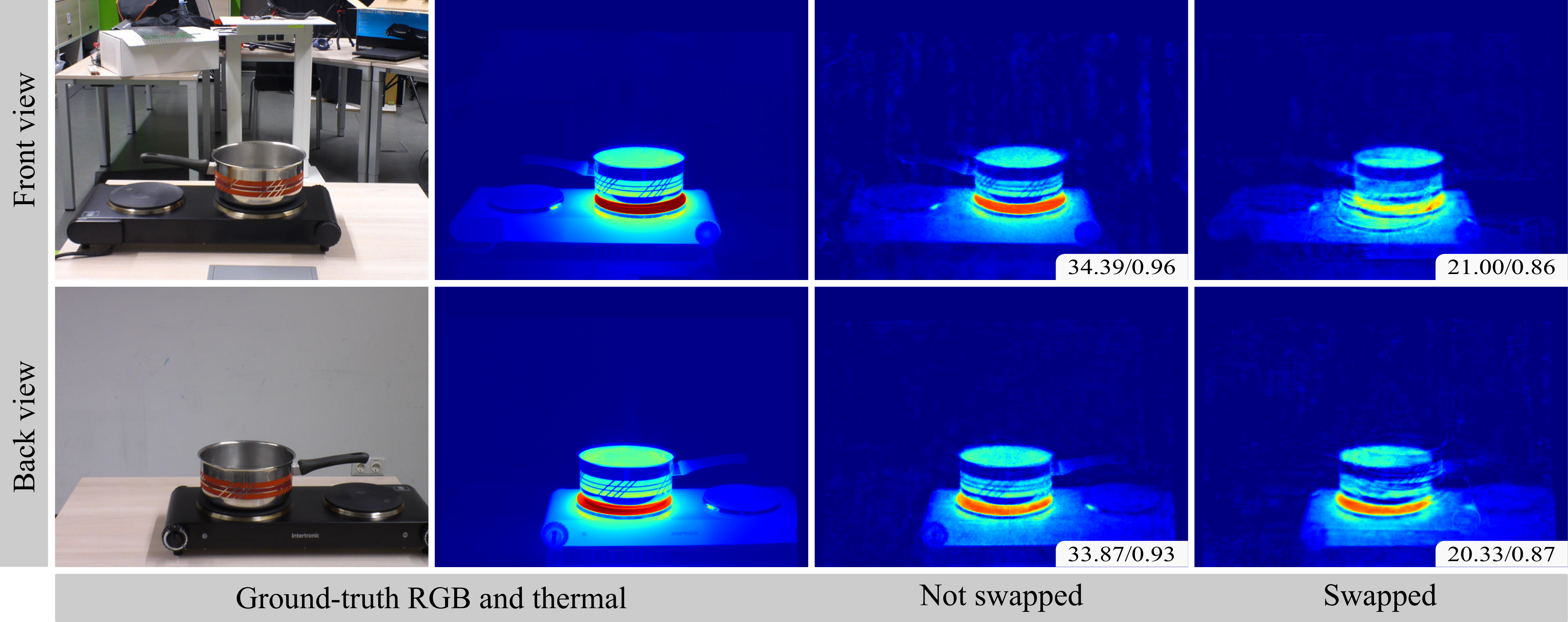}
    \caption{Training on 360-degree scenes with symmetrical objects, such as \textsc{Pan}, is challenging in the context of thermal imaging. In this experiment, we trained RGB-$t$ on two datasets: the original dataset, and a dataset in which we exchanged front and back views. We show a novel front-facing view in the first row and a back-facing view in the second row. We also report PSNR and SSIM (higher is better).}
    \label{fig:swapped_vs_notswapped}
\end{figure}

Another issue arises with objects that have symmetrical shapes, such as \textsc{Pan} in our work. Both, the front and back views of such objects display very similar backgrounds in thermal imaging due to its nature. This similarity in views, despite being 180 degrees apart in viewing direction, can cause ambiguities in training. To investigate this, we swapped the front and back views of the pan object and trained RGB-$t$ for it. We found that the results remained satisfactory, as demonstrated in Fig. \ref{fig:swapped_vs_notswapped}.

\begin{table}
 \caption{Quantitative results on (a) depth maps and (b) RGB images, measured using PSNR and SSIM (higher is better). Results were obtained from NeRFs trained on RGB images and depth maps. FT is left out since its RGB component is similar to TS.}
    \begin{subtable}{.54\linewidth}
      \centering
       \begin{adjustbox}{width=\linewidth}
        \begin{tabular}{lcc|cc|cc|cc}
            \toprule
             & \multicolumn{2}{c|}{TS} & \multicolumn{2}{c|}{FT} & \multicolumn{2}{c|}{RGB-D} & \multicolumn{2}{c}{SC} \\
             \cmidrule(lr){2-3} \cmidrule(lr){4-5} \cmidrule(lr){6-7} \cmidrule(lr){8-9}
             & PSNR $\uparrow$ & SSIM $\uparrow$ & PSNR $\uparrow$ & SSIM $\uparrow$ & PSNR $\uparrow$ & SSIM $\uparrow$ & PSNR $\uparrow$ & SSIM $\uparrow$ \\
            \midrule
            \textsc{Snail} & 33.38 & \textbf{0.94} & 30.56 & 0.92 & \textbf{33.51} & \textbf{0.94} & 33.07 & \textbf{0.94} \\
            \textsc{Bear} & \textbf{34.56} & \textbf{0.95} & 31.78 & 0.94 & 34.09 & 0.94 & 33.82 & 0.94 \\
            \textsc{Elephant} & \textbf{32.11} & \textbf{0.94} & 30.81 & 0.92 & 32.08 & 0.93 & 31.95 & \textbf{0.94} \\
            \midrule\midrule
            & \textbf{33.35} & \textbf{0.94} & 31.05 & 0.93 & 33.23 & \textbf{0.94} & 32.95 & \textbf{0.94} \\
            \bottomrule
        \end{tabular}
    \end{adjustbox}
    \caption{Depth reconstruction quality.}
    \end{subtable}\hfill
    \begin{subtable}{.42\linewidth}
      \centering
    \begin{adjustbox}{width=1.01\linewidth}
        \begin{tabular}{lcc|cc|cc}
            \toprule
             & \multicolumn{2}{c|}{TS} & \multicolumn{2}{c|}{RGB-D} & \multicolumn{2}{c}{SC} \\
             \cmidrule(lr){2-3} \cmidrule(lr){4-5} \cmidrule(lr){6-7}
             & PSNR $\uparrow$ & SSIM $\uparrow$ & PSNR $\uparrow$ & SSIM $\uparrow$ & PSNR $\uparrow$ & SSIM $\uparrow$ \\
            \midrule
            \textsc{Snail} & \textbf{39.94} & \textbf{0.97} & 36.05 & 0.96 & 37.66 & \textbf{0.97} \\
            \textsc{Bear} & \textbf{38.10} & \textbf{0.96} & 35.60 & 0.94 & 37.78 & \textbf{0.96} \\
            \textsc{Elephant} & \textbf{39.71} & \textbf{0.97} & 36.86 & 0.95 & 38.39 & \textbf{0.97} \\
            \midrule\midrule
             & \textbf{39.25} & \textbf{0.97} & 36.17 & 0.95 & 37.94 & \textbf{0.97} \\
            \bottomrule
        \end{tabular}
    \end{adjustbox}
    \caption{RGB reconstruction quality.}
    \end{subtable} 
    \label{tab:results_depth}
\end{table}

\vspace{-0.9cm}

\section{Results on RGB and Depth Maps}
\label{sec:results_depth}
Finally, we present results obtained from multi-modal NeRFs trained on RGB and depth maps for all of the four strategies. 
We follow the same protocol as described in the main paper, see Section \ref{subsec:results_other}.

\begin{figure}
    \centering
    \includegraphics[width=\textwidth]{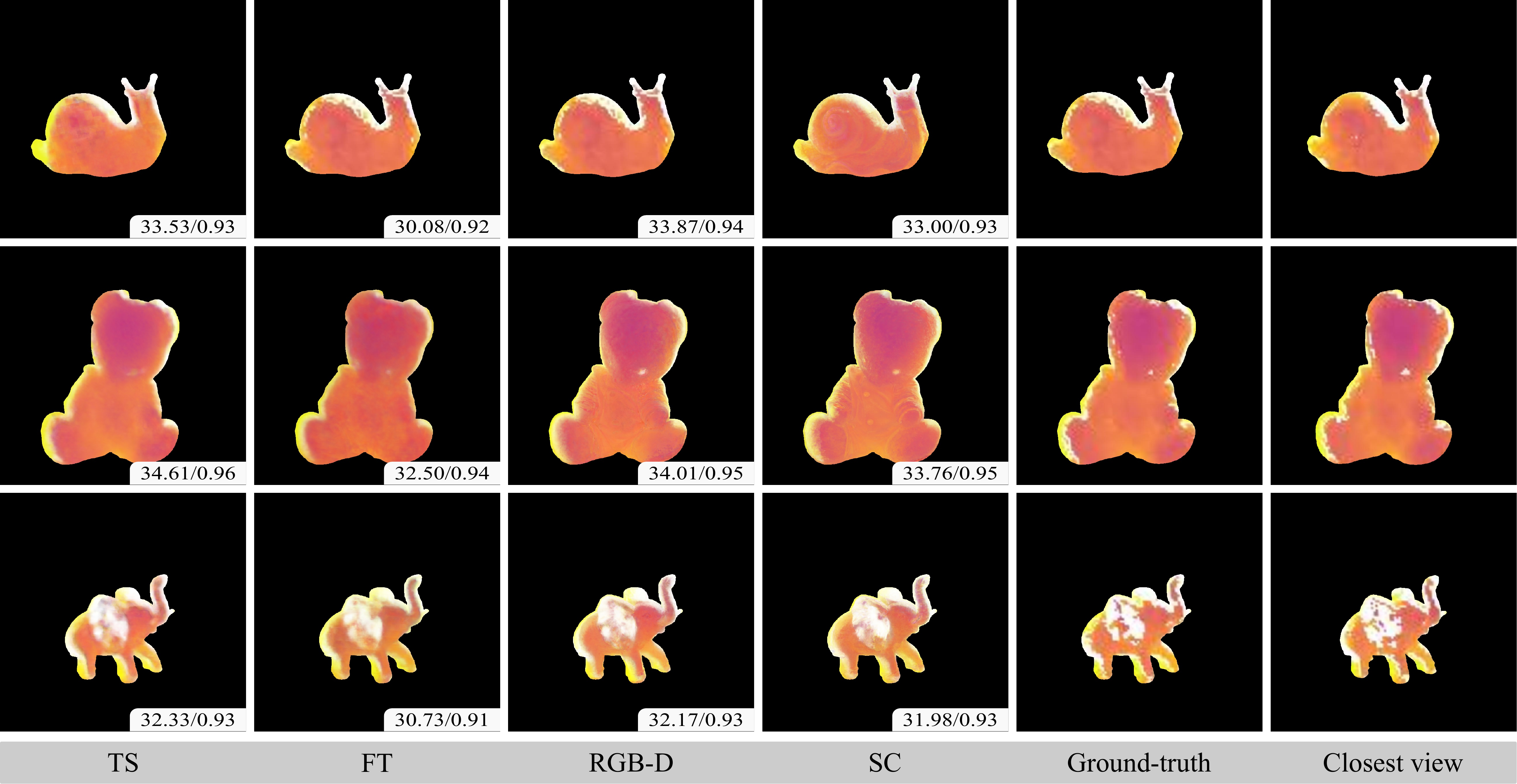}
    \caption{Reconstructions of a (left-out) depth map from multi-modal neural scene representations trained on RGB images and depth maps, arising from the four strategies that we compare. For each view, we also report PSNR and SSIM (higher is better). \textit{Closest view} denotes the nearest image in the training set.}
    \label{fig:results_rgb_d_d}
\end{figure}

\begin{figure}
    \centering
    \includegraphics[width=\textwidth]{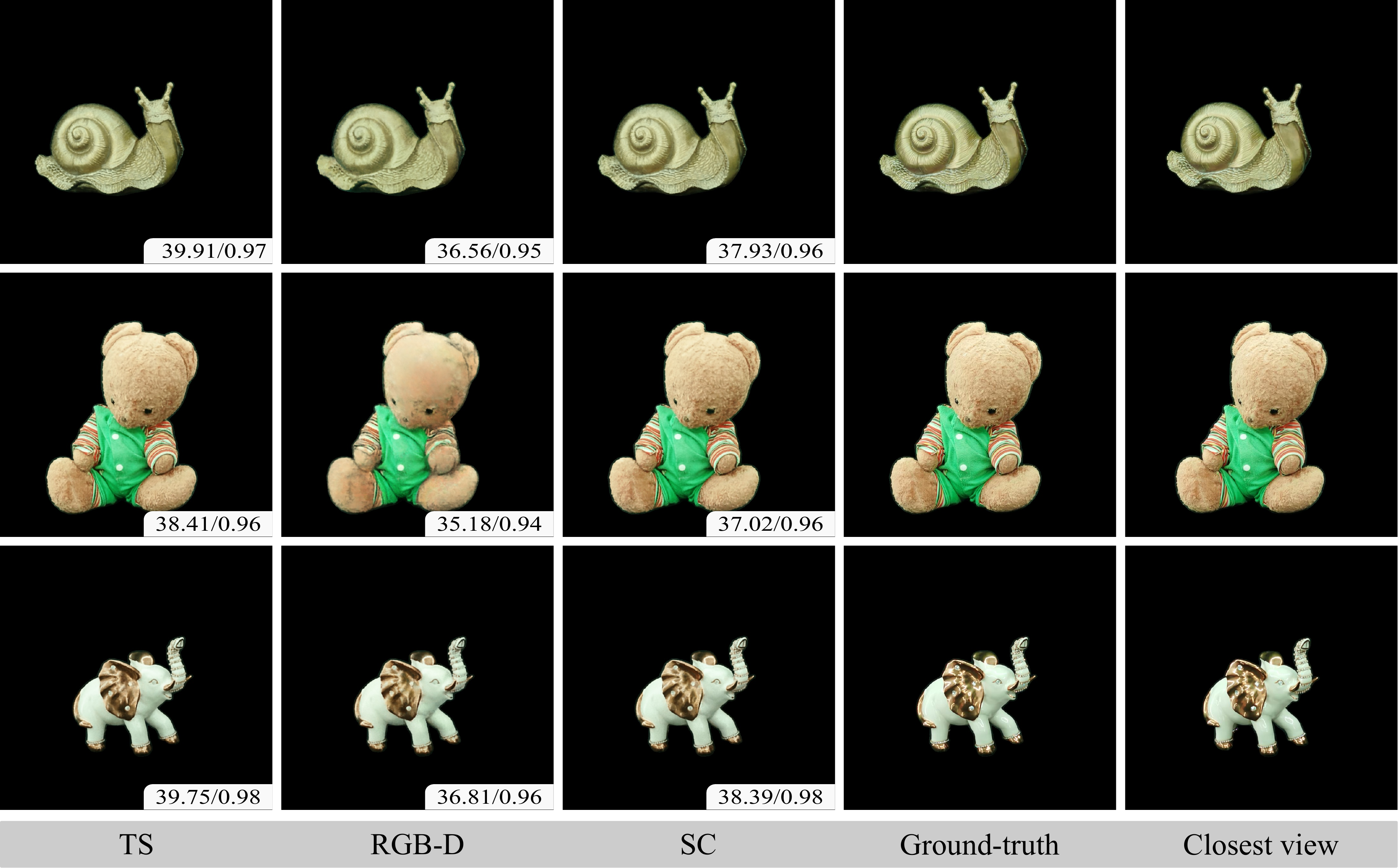}
    \caption{Reconstructions of a (left-out) RGB image from multi-modal neural scene representations trained on RGB images and depth maps, arising from the four strategies that we compare. FT is left out since its RGB component is similar to TS. For each view, we also report PSNR and SSIM (higher is better). \textit{Closest view} denotes the nearest image in the training set.}
    \label{fig:results_rgb_d_rgb}
\end{figure}

Quantitative and qualitative results can be found in Table \ref{tab:results_depth} and Figs. \ref{fig:results_rgb_d_d} and \ref{fig:results_rgb_d_rgb}.
Note that those findings match what we have seen in other modalities.
Specifically, considering depth reconstruction quality, we observe that TS and RGB-X (RGB-D in this case) perform almost on par, exhibiting similar behavior as for NeRFs trained on RGB and NIR images.
The same is true for RGB reconstruction quality; here, TS performs best, followed by SC and RGB-D.

\end{document}